\RequirePackage{fix-cm}
\documentclass[twocolumn,natbib]{svjour3}          % twocolumn
\smartqed  % flush right qed marks, e.g. at end of proof
\usepackage{graphicx}

\usepackage{amsmath}
\usepackage{amsfonts}
\usepackage{enumerate}
\usepackage[accsupp]{axessibility}

\usepackage{natbib}

\usepackage{ragged2e}
\usepackage{appendix}
\usepackage{booktabs}
\usepackage{color}
\usepackage{multirow}
\usepackage{multicol}
\usepackage{bm}
\usepackage{bbm}
\usepackage{caption}
\usepackage[labelformat=simple]{subcaption}
\usepackage{hyperref}
\hypersetup{
    colorlinks=true,
    linkcolor=blue,
    filecolor=blue,      
    urlcolor=blue,
    citecolor=cyan,
}

\usepackage{epstopdf}
\usepackage{amssymb}
\usepackage{enumitem}
\usepackage{color, colortbl}
\usepackage{wrapfig,lipsum}
\usepackage[dvipsnames,svgnames]{xcolor}
\usepackage{dsfont}
\usepackage{subcaption}
\usepackage[algo2e]{algorithm2e}
\usepackage{algorithm}
\usepackage{algorithmic}
\usepackage{amsmath}
\usepackage{caption} 
\usepackage[export]{adjustbox}% for the second solution

\usepackage{duckuments}% for the images
\usepackage{siunitx}
\usepackage{xcolor}
\usepackage{verbatim}
\usepackage[framemethod=tikz]{mdframed}

\usepackage{tabularx}
\newcolumntype{Y}{>{\centering\arraybackslash}X}
\usepackage{xspace}

\usepackage{times}
\usepackage{epsfig}
\usepackage{comment}
\usepackage{stfloats}
\newcolumntype{H}{>{\setbox0=\hbox\bgroup}c<{\egroup}@{}}
\usepackage{accents}
\newlength{\dhatheight}

\definecolor{Gray}{gray}{0.9}

\usepackage{ifpdf}

\usepackage{marvosym}

\begin{document}

\title{OPONeRF: One-Point-One NeRF for Robust Neural Rendering%\thanks{Grants or other notes
}
% \subtitle{Do you have a subtitle?\\ If so, write it here}


\author{Yu Zheng$^{1}$, 
        Yueqi Duan$^{2*}$,
        Kangfu Zheng$^{3}$,
        Hongru Yan$^{4}$,
        Jiwen Lu$^{1}$, 
        Jie Zhou$^{1}$
}

%\authorrunning{Dingfeng Shi, et al.} % if too long for running head

\institute{$^{*}$: Corresponding author\\
Yu Zheng$^{1}$\\
Email: yu-zheng@mail.tsinghua.edu.cn\\
Yueqi Duan$^{{2}}$\\
Email: duanyueqi@tsinghua.edu.cn\\
Kangfu Zheng$^{3}$\\
Email: zkf20@mails.tsinghua.edu.cn\\
Hongru Yan$^{4}$\\
Email: yanhr21@tsinghua.edu.cn\\
Jiwen Lu$^{1}$ \\
Email: lujiwen@tsinghua.edu.cn\\
Jie zhou$^{1}$ \\
Email: jzhou@tsinghua.edu.cn\\
\textsuperscript{1} Department of Automation, Tsinghua University. \\
\textsuperscript{2} Department of Electrical Engineering, Tsinghua University. \\
\textsuperscript{3} Department of Computer Science, Tsinghua University. \\
\textsuperscript{4} Weiyang College, Tsinghua University. \\
}

\date{Received: date / Accepted: date}

\maketitle

\begin{abstract}
In this paper, we propose a One-Point-One NeRF (OPONeRF) framework for robust scene rendering. 
Existing NeRFs are designed based on a key assumption that the target scene remains unchanged between the training and test time. 
However, small but unpredictable perturbations such as object movements, light changes and data contaminations broadly exist in real-life 3D scenes, which lead to significantly defective or failed rendering results even for the recent state-of-the-art generalizable methods. 
To address this, we propose a divide-and-conquer framework in OPONeRF that adaptively responds to local scene variations via personalizing appropriate point-wise parameters, instead of fitting a single set of NeRF parameters that are inactive to test-time unseen changes. 
Moreover, to explicitly capture the local uncertainty, we decompose the point representation into deterministic mapping and probabilistic inference. 
In this way, OPONeRF learns the sharable invariance and unsupervisedly models the unexpected scene variations between the training and testing scenes. % replace ``model'' 
To validate the effectiveness of the proposed method, we construct benchmarks from both realistic and synthetic data with diverse test-time perturbations including foreground motions, illumination variations and multi-modality noises, which are more challenging than conventional generalization and temporal reconstruction benchmarks. 
Experimental results show that our OPONeRF outperforms state-of-the-art NeRFs on various evaluation metrics through benchmark experiments and cross-scene evaluations. 
We further show the efficacy of the proposed method 
via experimenting on other existing generalization-based benchmarks and incorporating the idea of One-Point-One NeRF into other advanced baseline methods. 
Project Page: \url{https://yzheng97.github.io/OPONeRF/}. 
\end{abstract}

\section{Introduction}
Novel View Synthesis (NVS) is a critical research topic in computer vision and graphics, which has been widely used in online goods previewing, content generation applications and  filmmaking of bullet-time visual effect, etc. 
Given a 3D scene represented by a collection of posed images, NVS aims at rendering the scene from a free viewpoint. 
Early NVS methods, such as Structure-from-Motion (SfM)~\citep{hartley2003multiple} and image-based rendering~\citep{debevec1998efficient,fitzgibbon2005image,hedman2016scalable}, exploit geometric features from multi-view images which are time-consuming and usually require dense inputs to produce high-fidelity results. 

Recently, NVS methods equipped with the powerful deep-learning framework~\citep{mildenhall2020nerf,sitzmann2019scene,mildenhall2019local,lombardi2019nv} have raised extensive attention. 
In particular, Neural Radiance Fields (NeRFs)~\citep{mildenhall2020nerf,barron2021mip}  impressively demonstrate photorealistic results by encoding the 3D scene into a neural renderer implicitly. 
NeRFs learn a continuous volumetric function which is parameterized by a vanilla Multi-Layer Perceptron (MLP), mapping an querying 3D coordinate and 2D viewing direction into the volume density and view-dependent radiance of the sampled coordinate in the scene. 

\begin{figure*}[bt!]
	\centering
	\includegraphics[width=\textwidth]{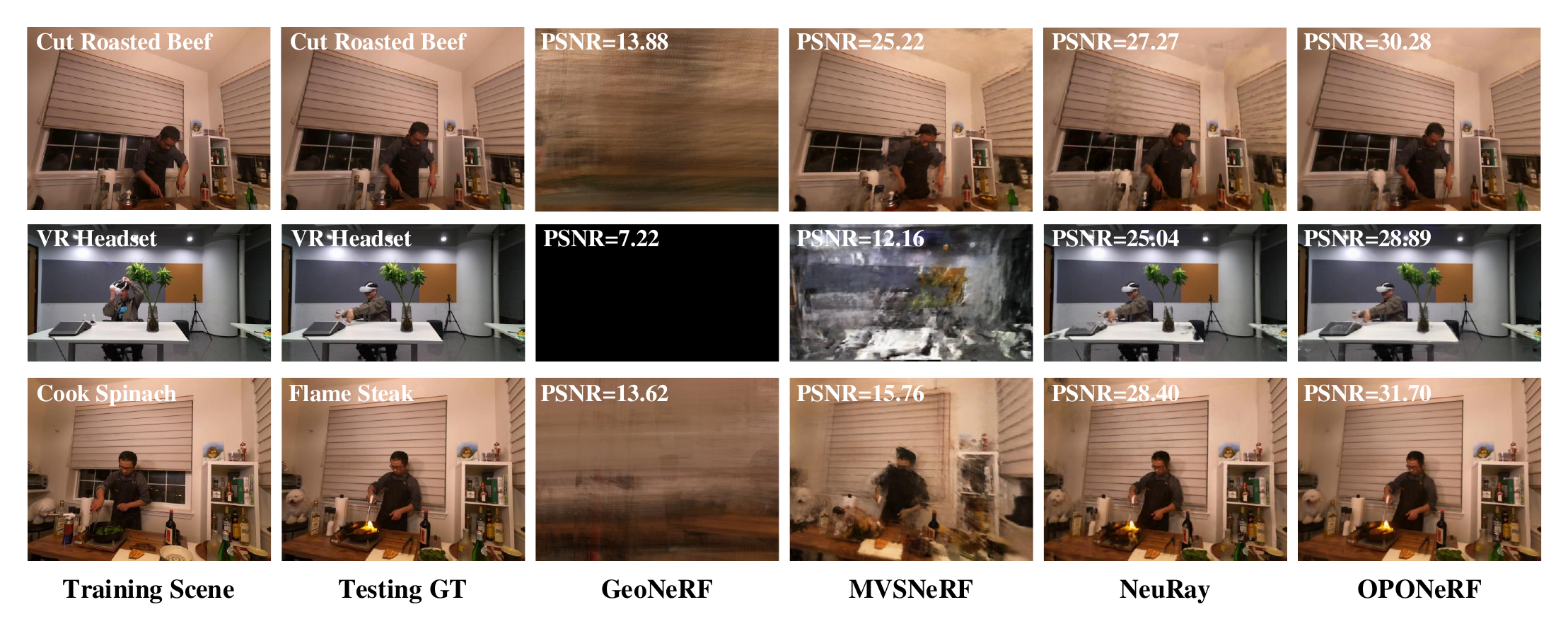}
	\vspace{-10mm}
	\caption{
		Qualitative comparison with baseline methods on the benchmarks of test-time scene perturbations. 
		The training scene before perturbation and testing groundtruth (GT) after perturbation belonging to the same camera view are shown. 
		In the top two rows, we trained on the first timeframe and tested on different timeframes of the same task scene. 
		In the bottom row, we trained on the first timeframe and tested on the timeframes of another task scene. 
		Advanced baselines such as MVSNeRF, NeuRay and GeoNeRF produce significantly defective rendering results or even fail when handling small but unpredictable test-time scene perturbations. 
		Notably, GeoNeRF renders all-black results when encountered with foreground movements. 
		OPONeRF better preserves the details on both moving foreground and static background areas. 
	} \label{fig:qualitative_main}
\end{figure*}

Despite rendering photorealistic results, existing NeRFs are designed based on a key assumption that the rendering target stays perfectly still throughout the training and testing phases, hence overfitting to a single static scene~\citep{trevithick2021grf}. 
However, in real-life 3D scenes, the assumption does not hold because perturbations such as foreground movements, illumination variations and data contaminations widely exist. 
A straightforward solution to these test-time scene changes is to apply generalizable methods~\citep{yu2021pixelnerf,wang2021ibrnet,trevithick2021grf,chen2021mvsnerf,liu2022neuray,johari2022geonerf,xu2022point} which combine the NeRF-like model with scene geometry. 
In this way, they condition the neural renderer on scene priors such as convolutional image features~\citep{trevithick2021grf,yu2021pixelnerf,wang2021ibrnet,johari2022geonerf}, cost volumes~\citep{chen2021mvsnerf,liu2022neuray} and attentional activation maps~\citep{wang2022attention}. 
While it only costs thousands of fine-tuning iterations for generalizable NeRFs to quickly generalize to unseen novel scenes~\citep{wang2021ibrnet,liu2022neuray}, we surprisingly observe that they poorly tackle such small but unpredictable test-time perturbations on the original rendering target. 
As shown in Fig.~\ref{fig:qualitative_main}, advanced generalizable NeRFs~\citep{chen2021mvsnerf,liu2022neuray,johari2022geonerf} produce significantly defective rendering results or even fail when addressing unexpected test-time scene changes. 
Moreover, although scene variations can be modeled temporally via spatio-temporal reconstruction~\citep{pumarola2021d,gao2021dynamic}, it is infeasible to learn the motion patterns given only a single timeframe at the training phase.  

In this paper, we propose a One-Point-One NeRF (OPONeRF) method for robust neural rendering. 
Instead of fitting all training points to a single set of parameters that are inactive to test-time changes, we design a divide-and-conquer framework that responds to local scene variations via personalizing appropriate NeRF parameters for each querying point. 
Specifically, OPONeRF subdivides the input scene into dense querying points produced by ray casting, for which we devise a parameter candidate decoder (PCD) module which decodes the rich geometric information into geometry-aware and layer-variant weight candidates. 
To personalize a neural renderer properly for the given point, OPONeRF learns an adaptiveness factor discriminative to its local geometry and responsible for employing appropriate point-wise renderer parameters from the candidate pool. 
Attributed to the abundant sampled points in the training scene, OPONeRF is provided with sufficient training samples so as to generalize and respond to unseen test-time feature points. 

To explicitly model the local uncertainty in a 3D scene, we further propose to represent the sampled point as the combination of deterministic invariance and probabilistic variances. 
Specifically, apart from the PCD module which determinisitcally maps the sampled point into the parameters of the neural renderer, we additionally correlate the point representation to multivariate Gaussian, where the approximation losses of variational parametric distribution are optimized. 
In this way, scene variations can be sampled from the learned distribution and thus expanding the expressiveness and flexibility of OPONeRF parameters. 
The probabilistic point representation is finally consumed by its personalized neural renderer and post-processed with volume rendering. 
To validate the effectiveness of the proposed method, we construct benchmarks from both real-life data~\citep{li2022neural,li2022streaming} and synthetic engines~\citep{greff2022kubric} with test-time scene changes, including foreground movements, illumination changes and exerted multi-model noises. 
Compared with conventional generalization and temporal reconstruction experimental settings, the new benchmark is much more challenging and better simulates the real-life scenario. 
While advanced generalizable NeRFs produce significantly defective results or even fail when confronted with those test-time scene changes, the proposed OPONeRF presents state-of-the-art rendering performance and can consistently improve the baseline methods. 
We also show that the proposed method is still applicable to the conventional generalization-based benchmarks where OPONeRF demonstrates competitiveness on the widely-used NeRF Synthetic~\citep{mildenhall2020nerf}, DTU~\citep{jensen2014large}, LLFF~\citep{mildenhall2019local} and Dynamic Scene~\citep{gao2021dynamic} datasets.

The contributions of this work are summarized as follows:
\begin{itemize}
	\item We present One-Point-One NeRF (OPONeRF) which learns a divide-and-conquer framework for robust scene rendering. 
	To handle the unexpected scene changes, OPONeRF learns the response to local scene variation by personalizing appropriate point-wise renderer parameters, so that it can adapt to unseen feature points at test time. 
	\item We propose to decompose the point-wise representation into deterministic mapping and probabilistic inference to explicitly capture the local uncertainty, so that the unexpected scene variations can be unsupervisedly modeled. 
	\item We construct new benchmarks from both realistic and synthetic data to validate our proposed method, which contains diverse test-time scene changes and are much more challenging than the conventional generalization and temporal reconstruction benchmarks. 
	\item We conduct extensive experiments on the proposed benchmarks and the results show that OPONeRF outperforms the state-of-the-art rendering methods and can consistently improve the baselines. We also show that OPONeRF is competitive on the conventional generalization-based benchmarks. 
\end{itemize}

\section{Related Work}
In this section, we briefly review the research topics on: 1) novel view synthesis, 2) neural radiance field and 3) uncertainty modeling in 3D perception. 

\subsection{Novel View Synthesis}
Novel view synthesis (NVS)~\citep{zhou2023single} is a long-standing topic studied in computer vision and graphics communities. 
The initial researches on NVS arose from image morphing~\citep{chen1993view}, light field rendering~\citep{levoy1996light} and multi-view stereo reconstruction~\citep{woodford2007new}. 
The deep networks enable the learning-based NVS methods. 
These works can be categorized into image-based rendering~\citep{flynn2016deepstereo,zhou2018stereo,riegler2020free} (IBR), volumetric representations~\citep{lombardi2019nv,liu2020neural,sitzmann2019deepvoxels}, neural scene encoding~\citep{sitzmann2019scene,mildenhall2020nerf,niemeyer2020differentiable,tancik2020fourier} and Gaussian Splatting based methods~\citep{kerbl20233d,wu20244dgaussians}. 

\subsection{Neural Radiane Fields}
Neural Radiance Fields (NeRFs) encode a 3D scene as a continuous volumetric radiance field of color and density. 
The vanilla NeRF~\citep{mildenhall2020nerf} consists of stacked multi-layer perceptrons (MLPs) and maps the querying coordinate and view direction to the corresponding color and density. 
Building on \citep{mildenhall2020nerf}, a number of works extend the usage of NeRF into other fields, including low-level tasks such as editing~\citep{liu2021editing,lan2024correspondence}, depth estimation~\citep{wei2021nerfingmvs,li2021mine} and pose estimation~\citep{yen2021inerf,jain2024learning}, and high-level tasks such as semantic segmentation~\citep{zhi2021place}, object detection~\citep{xu2023nerf} and classification~\citep{jeong2022perfception}. 
%Efforts are also made towards the improvements upon vanilla NeRF. 
%Plenoxels~\citep{fridovich2022plenoxels} significantly accelerate the convergence by converting the learning of neural networks into the optimization of spherical harmonic coefficients. 
NeRF also demonstrates its potential as a pre-training task useful for downstream tasks such as robotic object manipulation~\citep{driess2022reinforcement}. 
For a broader acquaintance with the research field on NeRF, we recommend the readers refer to the thorough literature review~\citep{gao2022nerf}. 
Our proposed method mainly focuses on the adaptiveness and robustness of NeRF to the given scene. 
While the adaptive ray sampling is proposed in \citep{kurz-adanerf2022}, it is designed to accelerate the rendering and achieve compact implicit representation. 
RawNeRF~\citep{mildenhall2022nerf} addresses the robustness to noises in both training and testing data inherently captured by raw sensors, while our OPONeRF handles the test-time scene changes compared with training data, including test-time noises. 

\textbf{NeRF Generalization: }
The vanilla NeRF~\citep{mildenhall2020nerf} requires dense posed inputs to overfit the scene and lacks the prior to generalize to novel scenes.  
%To learn scene prior from the training data, 
Efforts have been made towards generalizable NeRFs ~\citep{yu2021pixelnerf,trevithick2021grf,wang2021ibrnet,chen2021mvsnerf,johari2022geonerf,liu2022neuray,wang2022attention}. 
With strong generalization ability, finetuning on the novel scene is largely accelerated or even discarded by these methods. 
For example, GRF~\citep{trevithick2021grf}, PixelNeRF~\citep{yu2021pixelnerf} and IBRNet~\citep{wang2021ibrnet} use convolutional neural networks (CNN) to extract 2D features and retrieve them by re-projecting the 3D querying coordinate onto support images. 
%IBRNet~\citep{wang2021ibrnet} additionally incorporates a ray transformer network to aggregate all point-wise features along a sampled ray. 
%MVSNeRF~\citep{chen2021mvsnerf} extracts 3D geometric-aware features which demonstrate favourable effectiveness and generalizability. 
MVSNeRF~\citep{chen2021mvsnerf} builds cost volumes using warping transformation.  %which are then interpolated into the positions of querying points. 
%GeoNeRF~\citep{johari2022geonerf} extract strong geometric features and adopts attention mechanism to aggregate the point-wise features along a sampled ray. 
GeoNeRF~\citep{johari2022geonerf} and GNT~\citep{wang2022attention} adopt attention mechanism to aggregate the point-wise features along a sampled ray. 
NeuRay~\citep{liu2022neuray} proposes the occlusion-aware descriptors to handle the information inconsistency from different views. 
%NeRFusion~\citep{zhang2022nerfusion} sequentially predicts the local radiance field and incrementally constructs a global representation through a recurrent neural network. 
%To enhance the point-wise features projected from the image planes, 
GPNR~\citep{suhail2022generalizable} proposes a 3-stage Transformer architecture operating on local patches and encoding sampling positions in a canonical space. 
Inspired by MipNeRF~\citep{barron2021mip}, LIRF~\citep{huang2023local} tackles the feature inaccuracy and supervision ambiguity via learning a local implicit ray function which enable representing continuous sampling space in conical frustums. 
Point-NeRF~\citep{xu2022point} further incorporates the dense depth maps and aggregates them into point cloud features in the 3D space. 
These methods learn a fixed renderer with impressive generalization ability from the training prior. 
Our OPONeRF addresses the robustness to scene changes at test time, which is also critical in the field whereas the topic is missed by existing researches. 
Note that our OPONeRF is still applicable to benchmarks on NeRF generalization, which is a special case of test-time changes. 

\textbf{NeRF Benchmarks: }
NeRF benchmarks have recently been constructed with various application scenarios. 
For example, they evaluate NeRFs from the aspects of rendering quality~\citep{mildenhall2020nerf,barron2021mip}, training and testing efficiency~\citep{hu2022efficientnerf,reiser2021kilonerf}, editting controllability~\citep{yuan2022nerf}, scability~\citep{tancik2022block,turki2022mega} or applicability to downstream tasks~\citep{zhi2021place,jeong2022perfception,liu2023semantic,xu2023nerf}. 
More relevant to our benchmark are the generalization and spatio-temporal reconstruction benchmarks, as shown in Fig.~\ref{fig:setting_comparison}. 
In the generalization setting~\citep{trevithick2021grf,chen2021mvsnerf,liu2022neuray,johari2022geonerf,wang2022attention}, the training phase has already seen massive scene variations, thereby benefitting the fast generalization to test-time unseen scenes within a few fine-tuning iterations. 
In spatio-temporal reconstructions, multiple timeframes are provided to model the additional temporal dimension via NeRF+t~\citep{gao2021dynamic} or neural ODE solver~\citep{tian2023mononerf}. 
In contrast, only a single timeframe is provided during training phase in our proposed benchmark, which is much more challenging for generalizable methods~\citep{chen2021mvsnerf,liu2022neuray,wang2021ibrnet,johari2022geonerf,wang2022attention} and even infeasible for temporal modeling~\citep{gao2021dynamic,tian2023mononerf}. In \citep{wang2023benchmarking}, the authors also present a benchmark with image blurring, noises, fogging and pixelate to measure the robustness of NeRFs. 
Such data corruptions are exerted at both training and testing stages in \citep{wang2023benchmarking} while our proposed benchmark includes a different scenario, i.e., test-time data contaminations.

\subsection{Uncertainty Modeling in 3D Perception}
The idea of uncertainty modeling is popularly employed in 3D computer vision to handle data scarcity and ill-posedness. 
For instance, \citep{lu2021geometry} estimates depths as probability distribution to provide more accurate clues for monocular 3D detection. 
The pairwise positional relation between objects are estimated using probabilistic voting~\citep{du2020spot} in complex, cluttered indoor scenes. 
For denser prediction tasks, labels with uncertainty can be propagated from their neighborhood to eliminate semantic shifts~\citep{yang2023geometry}. 
In the field of neural radiance field, ActiveNeRF~\citep{pan2022activenerf} regards uncertainty as the optimizing target under constrained input budget. 
Stochastic-NeRF~\citep{shen2021stochastic} and NeRF-VAE~\citep{kosiorek2021nerf} also take the variational model into consideration. 
The former models the static appearance properties such as the radiance-density pairs as stochastic variables. 
The latter follows a similar pattern to generalization-based NeRFs that condition the whole set of fixed parameters on a latent code. 
In contrast, our OPONeRF targets on the sampled point itself which subdivides the overall scene and discriminatively responds to local variations via personalizing point-wise parameters, thereby representing the scene with more expressiveness and flexibility. 
\begin{figure*}[tb]
	\centering
	\includegraphics[width=0.98\textwidth]{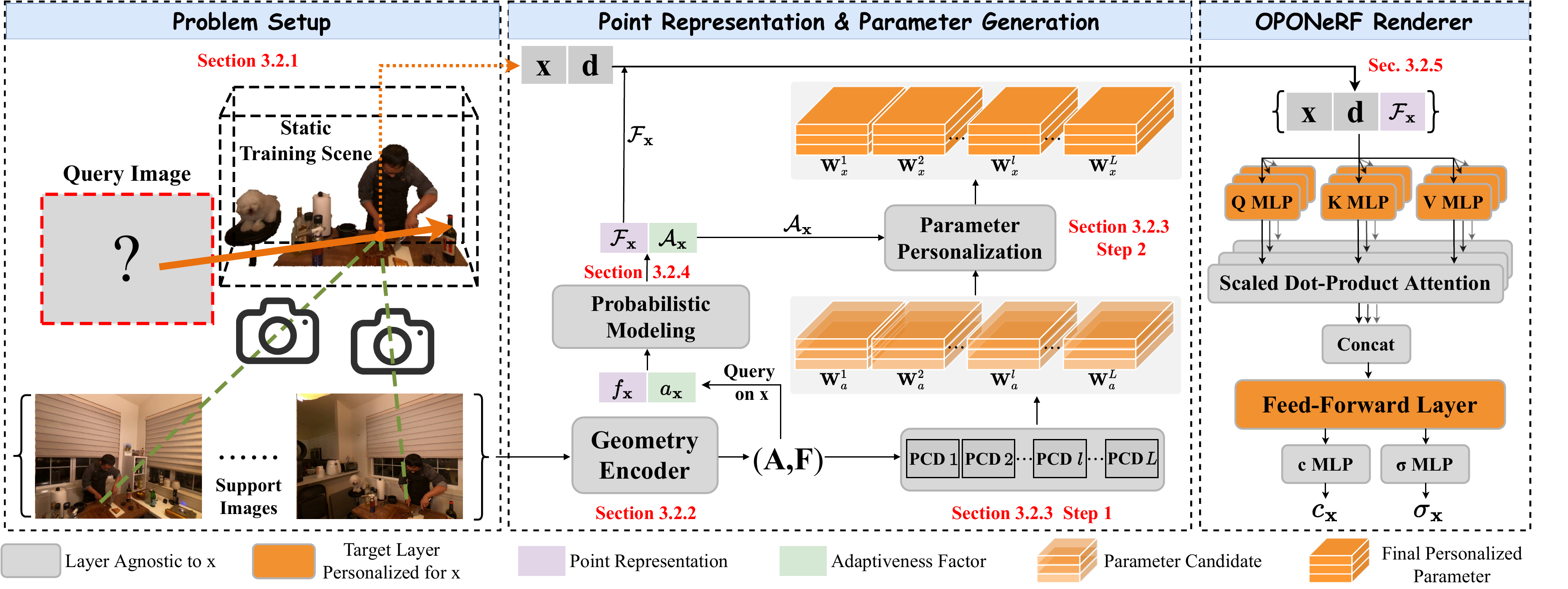}
	\vspace{-3mm}
	\caption{
		Illustration of OPONeRF. 
		To render the querying view, OPONeRF firstly extracts a pair of scene-level features ($\mathbf{F}$ and $\mathbf{A}$) using a geometry encoder. 
		The initial point representation $f_\mathbf{x}$ and adaptiveness factor $a_\mathbf{x}$ are interpolated from $\mathbf{F}$ and $\mathbf{A}$ via querying on $\mathbf{x}$. 
		OPONeRF then parallelly learns a series of PCDs taking $\mathbf{F}$ as input and producing the geometry-aware and layer-variant candidate parameters $\mathbf{W}^l_a$ for the target layers of OPONeRF renderer. 
		For each $\mathbf{x}$, we learn its final probabilistic representation $\mathcal{F}_\mathbf{x}$ and the fused $\mathcal{A}_\mathbf{x}$. 
		The renderer parameters personalized for each $\mathbf{x}$ are adaptively controlled by $\mathcal{A}_\mathbf{x}$ via selecting from candidate parameters. 
		In this way, OPONeRF learns a personalized neural renderer for each sampled $x$. 
		The OPONeRF renderer is a Ray Transformer with layers personalized for each $\mathbf{x}$ as well as shareable ones that are agnostic to $\mathbf{x}$. 
		The output of OPONeRF renderer will be processed by the classical volume rendering to get the properties of final querying view. 
		(\textit{Best viewed in color.})
	} \label{fig:opo_overall_pipeline}
\end{figure*}

\section{Approach}
In this section, we firstly review the preliminary on the neural representation in NeRFs which our method is based on (Sec.~\ref{sec:representation}). 
Then we elaborate on the framework of OPONeRF which consists of the overall representation (Sec.~\ref{sec:oponerf_representation}), the geometry encoder (Sec.~\ref{sec:geometry_encoder}) and the OPONeRF decoder (Sec.~\ref{sec:adaptiveness_decoder}) personalized for discriminative and probabilistic point representation (Sec.~\ref{sec:variational_point_representation}). 
Finally we present the rendering and optimizing target (Sec.~\ref{sec:rendering_oponerf}), implementation details (Sec.~\ref{sec:implementation_details}) and provide a more in-depth comparison with other  related methods (Sec.~\ref{sec:discussion}). 

\subsection{Preliminary on NeRF Representation}
\label{sec:representation}
A neural radiance field typically comprises the learning of a plenoptic mapping function $\mathcal{M}_\Theta$. 
Specifically, using known extrinsics and intrinsics of the camera, we can simulate a ray emitting from the camera pinhole with a two-dimensional viewing direction $\mathbf{d}=\{\theta, \phi\}$ and going through a pixel of the querying image. 
For a sampled 3-dimensional spatial coordinate $\mathbf{x}=\{x,y,z\}$ on $\mathbf{r}$, $\mathcal{M}_\mathrm{\Theta}$ learns a neural renderer with the mapping: 
\begin{equation}
	\label{eq:nerf_mapping}
	\mathcal{M}_\Theta: (\mathbf{x}, \mathbf{d}) \longrightarrow (\mathbf{c}, \sigma),
\end{equation}
where $\Theta$ denotes the network parameters of neural renderer in vanilla NeRFs~\citep{mildenhall2020nerf}  shared by all $\mathbf{x}$s. 
The scene properties $\mathbf{c}\in\mathbb{R}^3$ and $\sigma\in\mathbb{R}$ represent the radiance and density at $\mathbf{x}$ respectively. 
To avoid overfitting to the training scene, the generalization-based~\citep{trevithick2021grf,chen2021mvsnerf,liu2022neuray,johari2022geonerf,wang2021ibrnet,wang2022attention} NeRFs aim to capture scene prior by concatenating the geometry-agnostic $[\mathbf{x}, \mathbf{d}]$ input with the geometry-aware $f_\mathbf{x}$: 
\begin{equation}
	\label{eq:generalizable_nerf_mapping}
	\mathcal{M}_\Theta^{gen}: (\mathbf{x}, f_\mathbf{x}, \mathbf{d}) \longrightarrow (\mathbf{c}, \sigma),
\end{equation}
where $f_\mathbf{x}$ represents the point-wise feature at $\mathbf{x}$ aggregated from the support views. 
Similar to the mapping mechanism of vanilla NeRFs in Eqn.~(\ref{eq:nerf_mapping}), the NeRF parameters $\Theta$ are still shared by all sampled points and fixed at test time. 
We perform the physical and differentiable volume rendering~\citep{kajiya1984ray} as post-processing to obtain the pixel properties on the querying views.

\subsection{Proposed Method}
Our proposed OPONeRF aims to tackle the robust rendering under the scenario of test-time unseen local representations or scene perturbations. 
The overall framework is presented in Fig.~\ref{fig:opo_overall_pipeline}. 
Note that the framework in Fig.~\ref{fig:opo_overall_pipeline} is shared by all experiments on the robustness benchmark. 
\subsubsection{OPONeRF Representation}
\label{sec:oponerf_representation}
Compared with generalization-based NeRFs, our OPONeRF further conditions the parameters $\Theta$ of neural renderer on the representation of sampled point $\mathbf{x}$: 
\begin{equation}
	\label{eq:scene_adaptive_nerf_mapping}
	\mathcal{M}_\Theta^{opo}(\mathbf{x}): (\mathbf{x}, f_\mathbf{x}, \mathbf{d}, \mathbf{F}) \longrightarrow (\mathbf{c}, \sigma), 
\end{equation}
where $\mathbf{F}$ represents the overall representation of the scene extracted by a geometric encoder. 
OPONeRF dynamically conditions the parameters of plenoptic mapping on the sampled point to formulate the personalization of point-wise neural renderer. 

The overall network architecture in Fig.~\ref{fig:opo_overall_pipeline} contains a geometry encoder which extracts rich geometric features, a series of parallel parameter candidate decoders (PCD) which provide a geometric-aware and layer-variant parameter pool, and the final personalized neural renderer discriminative to each sampled point with probabilistic modeling. 

\subsubsection{Geometry Encoder}
\label{sec:geometry_encoder}
Given a set of posed support images $\{I_k\}_{k=1}^K$, one can encode rich geometric features using an encoder $E_g(\cdot)$ such as MVS stereo~\citep{chen2021mvsnerf} or projection-and-aggregation~\citep{trevithick2021grf,wang2021ibrnet}: 
\begin{equation}
	\label{eq:conventional_geometric_encoder}
	\mathbf{F} = E_g(\{I_k\}). 
\end{equation}
In practice we follow \citep{chen2021mvsnerf} to extract cost volumes which condense the appearance displacement between support views and one pre-set reference view ($k=1$)~\citep{yao2018mvsnet}: 

\begin{equation}
	\label{eq:cost_volume_to_F}
	\begin{aligned}
		%(u,v,z) &= \Pi_k (x,y,z) \\
		&F_k = T(I_k), \\
		&F_{k,z}(u,v) = W_k\bigl(\mathcal{H}_k(z)[u,v,1]^T\bigr), \\
		&\mathbf{F} = B\Bigl(\text{Var} \bigl(\{F_{k,z}(u,v)\}\bigr)\Bigr),
	\end{aligned}
\end{equation}

We set $T(\cdot)$ as a shared CNN that extracts 2D feature map $F_k$ from each support view. The cost volume is constructed by warping all $F_k$s to the canonical reference view and computing their view-wise variance. The homographic warping matrix from the $k$-th support view to the reference view at depth $z$ is denoted by $\mathcal{H}_k(z)$. The warped feature at pixel $(u,v)$ and depth $z$ in the reference view is represented by $F_{k,z}(u,v)$. The overall scene representation $\mathbf{F}$ is obtained by feeding the cost volume into a 3D CNN $B(\cdot)$. For a sampled point $\mathbf{x}$ in the continuous 3D space, its point-wise feature $f_\mathbf{x}$ can be interpolated from the gridded $\mathbf{F}$.

\subsubsection{Point-wisely Personalized OPONeRF Renderer}
\label{sec:adaptiveness_decoder}
Given the overall scene representation $\mathbf{F}$, a sampled point coordinate $\mathbf{x}$, its viewing direction $\mathbf{d}$ and point-wise feature $f_\mathbf{x}$, OPONeRF personalizes the parameters of $L$ targeted renderer layers for $\mathbf{x}$ in two steps, namely parameter candidate generation and discriminative parameter personalization. 
\begin{figure}[tb]
	\centering
	%\vspace{-2mm}
	\includegraphics[width=0.48\textwidth]{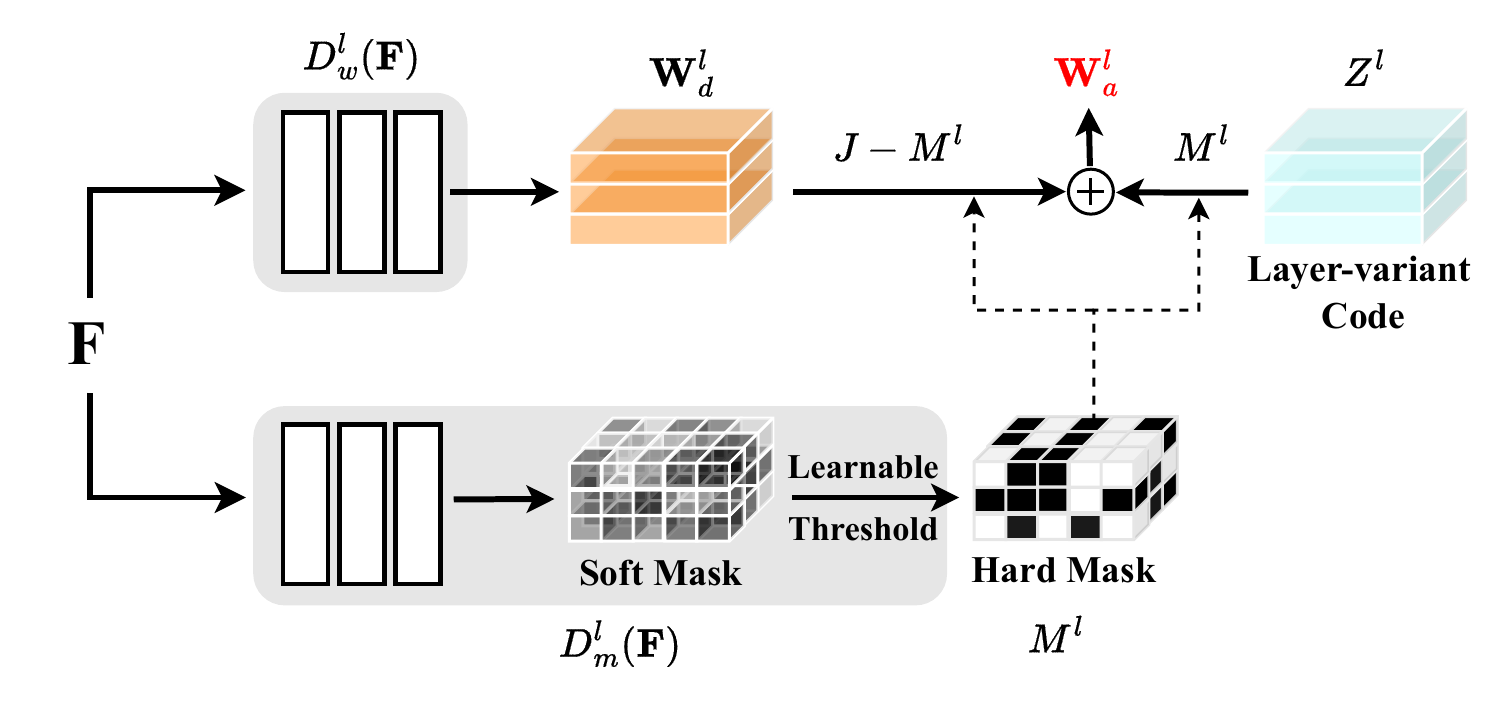}
	\vspace{-8mm}
	\caption{
		The parameter candidate decoder (PCD) responsible for the $l$-th target layer parameterized by $\mathbf{W}^l_a$. 
		OPONeRF parallelly learns a series of PCDs each responsible for a unique target layer. 
	} \label{fig:pcd_module}
\end{figure}

\textbf{Step 1. Parameter Candidates Generation: }
Suppose the $l$-th target layer in the OPONeRF renderer is parameterized by $\hat{\mathbf{W}}^l \in \mathbb{R}^{C_i^l\times C_o^l}$. 
We design a series of parallel parameter candidate decoders (PCD, shown in Fig.~\ref{fig:pcd_module}) to firstly decode $\mathbf{F}$ into weight candidates for the target layers: 
\begin{equation}
	\label{eq:parameter_candidate_decoder}
	\mathbf{W}^l_a = \text{PCD}^l(\mathbf{F}), \mathbf{W}^l_a\in \mathbb{R}^{C_i^l\times C_o^l}. 
\end{equation}
One PCD contains a paired $D^l_w(\cdot)$ and $D^l_m(\cdot)$ responsible for parameters and hard binary masks respectively. 
Specifically, for each target layer, we firstly decode $\mathbf{F}$ into a parameter matrix: 
\begin{equation}
	\label{eq:soft_mask_generation}
	\{\mathbf{W}^l_d|\mathbf{W}^l_d = D^l_w(\mathbf{F}),  l=1,2,...,L\}, 
\end{equation}
where $\mathbf{W}^l_d\in \mathbb{R}^{C_i^l\times C_o^l}$. 
To prevent $\mathbf{W}^l_d$ being trivially the same for all $L$ targeted layers, we additionally learn a set of layer-variant codes $\mathbf{Z}^l\in \mathbb{R}^{C_i^l\times C_o^l}$ which are fused with $\mathbf{W}^l_d$ using element-wise binary mask $M^l$: 
\begin{equation}
	\label{eq:soft_weight_generation}
	\begin{aligned}
		&\mathbf{W}^l_a = (J-M^l)*\mathbf{W}^l_d + M^l*\mathbf{Z}^l,  \\
		&\text{where} \; M^l = D^l_m(\mathbf{F}), 
	\end{aligned}
\end{equation}
where $*$ and $J$ denote the Hadamard product and all-ones matrix respectively. 
$\mathbf{Z}^l$s are randomly initialized independently and iteratively updated during training, providing a compact representation of each target layer. 
In this way, the outputs of PCDs are both geometry-aware and layer-variant. 
Note that instead of using a fixed scalar threshold, we rather additionally learn a threshold matrix element-wisely correspondent with $M^l$ for binary quantization, so that the parameter candidates can be learned with more flexibility. 

\textbf{Step 2. Discriminative Parameter Personalization: }
For each sampled $\mathbf{x}$, OPONeRF discriminatively personalizes its point-wise parameters of the target layers in the neural renderer. 
A naive attempt is to learn a totally unique renderer responsible for each 3D location. 
Suppose for an image with $h$ the height, $w$ the width and $N$ the number of sampled points on a ray, it requires $h\times w\times N$ set of learnable matrices for a single target layer which brings unacceptable computational and optimization overhead. 
Given this, we regard $\mathbf{W}^l_a$ as a parameter pool and select from it the proper ones via introducing an adaptiveness factor $\mathcal{A}_\mathbf{x}\in \mathbb{R}^L$ discriminative to each $\mathbf{x}$. 
$L$ is the number of target layers in the OPONeRF renderer. 
We detail the calculation of $\mathbf{A}_x$ in Sec.~\ref{sec:variational_point_representation}. 
Considering the large quantities of sampled points, $\mathcal{A}_\mathbf{x}$ learns how to personalize the neural renderer in a layer-wise manner rather than element-wise masks $M^l$. 
$\mathcal{A}_\mathbf{x}$ adaptively employs the parameter candidates in Eqn.~(\ref{eq:soft_weight_generation}): 
\begin{equation}
	\label{eq:layer_mask}
	\mathbf{W}^l_x = (J-\mathcal{A}_\mathbf{x}[l]J)\mathbf{W}^l_a + \mathcal{A}_\mathbf{x}[l]\mathcal{Z}^l, 
\end{equation}
where $\mathbf{W}^l_x$, $\mathcal{A}_\mathbf{x}[l]$, $\mathcal{Z}^l$ are the final parameters of the $l$-th target layer personalized for $\mathbf{x}$, $l$-th element of $\mathcal{A}_\mathbf{x}$ and another layer-variant code of the $l$-th target layer respectively. 
$\mathcal{Z}^l$ is randomly initialized and iteratively updated similar to $Z^l$. 
%$\mathbf{a}(x)$ can be regarded as a layer mask which is much coarser-grained than $M^l$. 
We do not apply threshold-based quantization to binarize $\mathcal{A}_\mathbf{x}$ but rather adopt soft weighted aggregation of $\mathbf{W}^l_a$ and $\mathcal{Z}^l$ in Eqn.~(\ref{eq:layer_mask}). 
Although we aim to personalize unique renderer parameters for each $\mathbf{x}$, Eqn.~(\ref{eq:layer_mask}) can be implemented with matrix multiplication (e.g., \texttt{torch.bmm}) for a batched points to maintain the computational efficiency, which we show in the experiment section. 

\begin{figure}[tb]
	\centering
	\includegraphics[width=0.48\textwidth]{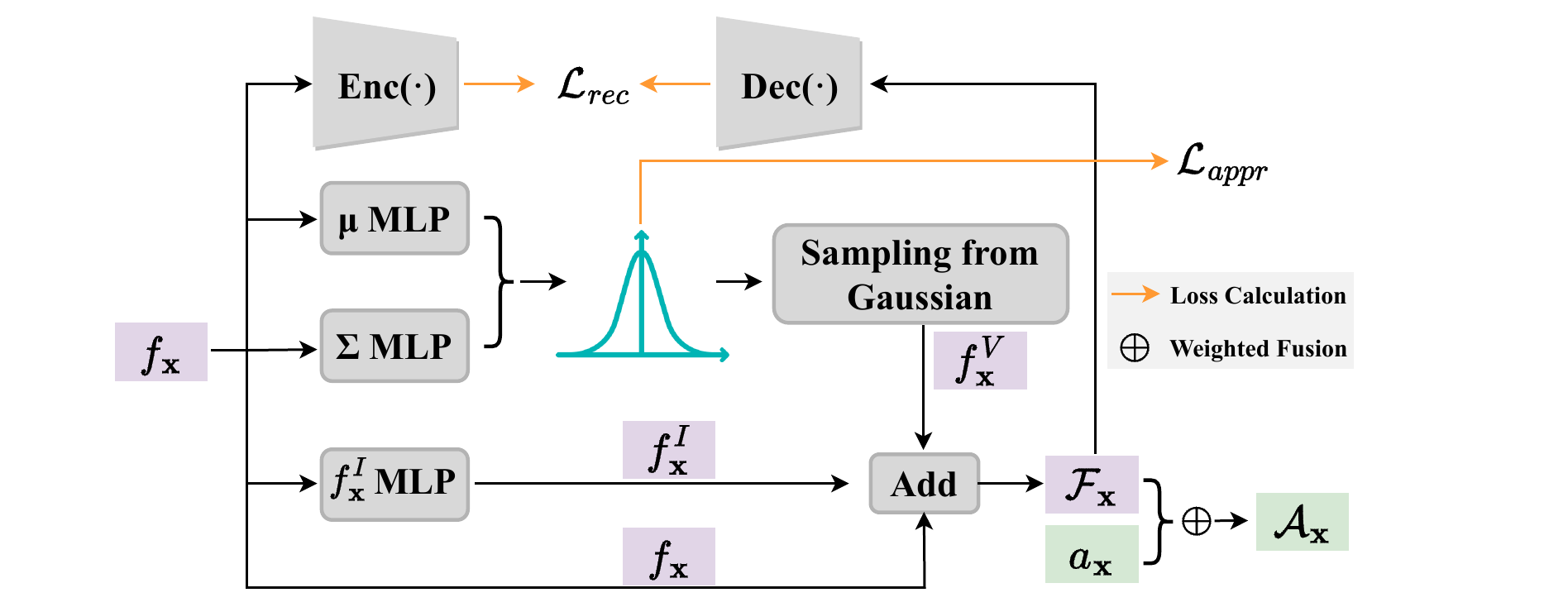}
	\vspace{-6mm}
	\caption{
		Probabilistic modeling of point representation. 
	} \label{fig:vae_modeling}
\end{figure}

\subsubsection{Probabilistic Modeling of Point Representation}
\label{sec:variational_point_representation}
With the deterministic scene representation, the mapping of ``One Point $\rightarrow$ One NeRF'' is constrained within the point feature space in the training scene. 
Considering the unpredictable nature of scene perturbation, we propose to explicitly model the point feature $f_\mathbf{x}$ with uncertainty, where the parameters of targeted layers can be flexibly sampled from a probabilistic feature space. 
%Specifically, we model the point representation as the combination of deterministic invariance and unexpected variance: $\mathcal{F}_\mathbf{x} \triangleq \alpha f_\mathbf{x}^I + \beta f_\mathbf{x}^V$. 
Specifically, we assume the scene representation at location $\mathbf{x}$ is correlated with a random process and can be modeled as the combination of deterministic invariance ($f_\mathbf{x}^I$) and unexpected variance ($f_\mathbf{x}^V$): 
\begin{equation}
	\label{eq:point_feature_decomposition}
	\mathcal{F}_\mathbf{x}  \propto \alpha f_\mathbf{x}^I + \beta f_\mathbf{x}^V. 
\end{equation}

We model its randomness by assuming $f_\mathbf{x}^V$ subject to a multivariate Gaussian prior $p_\theta^*(f_\mathbf{x}^V) = \mathcal{N}(f_\mathbf{x}^V;\mathbf{0},\mathbf{I})$. 
Since $p_\theta^*(f_\mathbf{x}^V)$ is usually intractable, in practice, we approximate its variational posterior $q_\phi$ with: 
\begin{equation}
	\label{eq:approximate_gaussian_of_point_feature}
	\text{ln}q_\phi(f_\mathbf{x}^V|f_\mathbf{x}) = \mathcal{N}\bigl(f_\mathbf{x}^V;\bm{\mu}_\mathbf{x},\bm{\Sigma}_\mathbf{x}^2\bigr), 
\end{equation}
where $\bm{\mu}_\mathbf{x}$ and diagonal $\bm{\Sigma}_\mathbf{x}$ are regressed by the corresponding MLP layers. 
Moreover, a generative model is responsible for approximating the likelihood $p_\theta(f_\mathbf{x}|f^V_\mathbf{x})$ by reconstructing the original point representation. 
To fit the abovementioned parameterized probability distribution, we jointly optimize the parameters of the generative model ($\theta$) and the approximated posterior ($\phi$) via minimizing the evidence lower bound (ELBO): 
\begin{equation}
	\label{eq:elbo_loss}
	\mathcal{L}_{\mathbf{\mu}_\mathbf{x}, \mathbf{\sigma}_\mathbf{x}}(f_\mathbf{x}) = \text{ln}p_\theta(f_\mathbf{x}|f_\mathbf{x}^V) - D_{KL}\bigl(q_\phi(f_\mathbf{x}^V|f_\mathbf{x})||p_\theta(f^V_\mathbf{x})\bigr). 
\end{equation}

After reparameterization~\citep{bengio2013representation}, $\mathcal{L}_{\mathbf{\mu}_\mathbf{x}, \mathbf{\sigma}_\mathbf{x}}(f_\mathbf{x})$ can reduce to: 
\begin{equation}
	\label{eq:elbo_loss_after_reparam}
	\begin{aligned}
		\mathcal{L}_{\mathbf{\mu}_\mathbf{x}, \mathbf{\sigma}_\mathbf{x}}(f_\mathbf{x}) &\approx \frac{1}{B}\sum\limits^B_{i=1}\text{ln}p_\theta(f_\mathbf{x^i}|f_\mathbf{x^i}^V) \\
		+ \frac{1}{2B}\sum\limits^B_{i=1}\sum\limits^D_{j=1}\Bigl(1&+\text{ln}\bigl((\bm{\sigma}^{(j)}_{\mathbf{x}_i})^2\bigr) - (\bm{\mu}^{(j)}_{\mathbf{x}_i})^2 - (\bm{\sigma}^{(j)}_{\mathbf{x}_i})^2\Bigr), 
	\end{aligned}
\end{equation}
which corresponds to a reconstruction loss $\mathcal{L}_{rec}$ and Gaussian approximation (KL divergence) loss $\mathcal{L}_{appr}$. 
$B$ and $D$ are number of sampled points per mini-batch and dimension of multivariate Gaussian. 
We implement $\mathcal{L}_{rec}$ as squared difference for simplicity: 
\begin{equation}
	\label{eq:reconstruction_loss_simplify}
	\mathcal{L}_{rec} = \frac{1}{B}\sum\limits^B_{i=1}||Enc(f_\mathbf{x^i})-Dec(\mathcal{F}_\mathbf{x^i})||_2. 
\end{equation}
where we take $\mathcal{F}_\mathbf{x^i}$ instead of $f^V_\mathbf{x^i}$ given the deterministic assumption of $f^I_\mathbf{x^i}$. 
We additionally learn a pair of encoder $Enc(\cdot)$ and decoder $Dec(\cdot)$ to map $f_\mathbf{x^i}$ and $\mathcal{F}_\mathbf{x^i}$ into the same latent space. 

As shown in Fig.~\ref{fig:vae_modeling}, we implement the final probabilistic point feature $\mathcal{F}_\mathbf{x}$ with a residual connection for steady training, i.e., $\mathcal{F}_\mathbf{x} := f_\mathbf{x} + \alpha(f^I_\mathbf{x} +  f^V_\mathbf{x})$. 
$\alpha$ is a weight hyper-parameter. 
Then we formulate the adaptiveness factor $\mathcal{A}_\mathbf{x}$ as weighted fusion: 
\begin{equation}
	\label{eq:adaptiveness_factor_fusion}
	\mathcal{A}_\mathbf{x} = \mathbf{a}_\mathbf{x} \oplus\mathcal{F}_\mathbf{x}, \text{where} \;\; \includegraphics[height=1cm,valign=c]{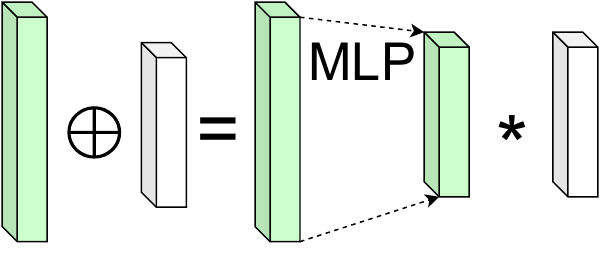}. 
\end{equation}

$\mathbf{a}_\mathbf{x}$ is computed by spatial interpolation similar to $f_\mathbf{x}$. 
Specifically, we modify the geometric encoder in Eqn.~(\ref{eq:conventional_geometric_encoder}) into: 
\begin{equation}
	\label{eq:ours_geometric_encoder}
	[\mathbf{F},\mathbf{A}] = E_g^{'}(\{I_v\}), 
\end{equation}
where the fan-out feature channel of the last layer in 3D CNN $B(\cdot)$ is extended by $L$. 
In concatenation with $\mathbf{F}$ on the feature dimension and aligned with $\mathbf{F}$ as gridded volumes on the spatial dimensions, $\mathbf{A}$ is then interpolated into $\mathbf{a}(\mathbf{x})$ values at different querying coordinates.

\subsubsection{Rendering of OPONeRF Renderer}
\label{sec:rendering_oponerf}
As shown in Fig.~\ref{fig:opo_overall_pipeline}, the final probabilistic point representation $\mathcal{F}_\mathbf{x}$ is consumed by the OPONeRF renderer, which consists of both targeted layers personalized for $\mathbf{x}$ and the other layers shared by all $\mathbf{x}$s. 
The renderer architecture is the Ray Transformer~\citep{wang2021ibrnet} where a sampled point attends to other points along the ray. 
The target layers are the query (Q), key (K), value (V) projection MLPs and feed-forward layer of Ray Transformer. 
We conduct the volume rendering in Sec.~\ref{sec:representation} to post-process the output of renderer to obtain the pixel values on the querying view. 
The volume rendering process is supervised by the photometric loss formulated by: 
\begin{equation}
	\label{eq:loss_photometric}
	\mathcal{L}_{pho} = \frac{1}{|R|}\sum\limits_{\mathbf{r}\in R}||\hat{\mathbf{C}}(\mathbf{r})-\mathbf{C}(\mathbf{r})||, 
\end{equation}
where $\mathbf{C}(\mathbf{r})$ and $R$ denote the groundtruth color of the pixel emitting the ray $\mathbf{r}$ and the set of rays in a mini-batch. 

%\textbf{Diversity Loss: }
To fully exploit the flexibility brought by the adaptiveness factors more thoroughly, we aim to moderately maintain the diversity of $\mathbf{a}(x)$. 
For $|R|$ rays each containing $N$ sampled points, we design a diversity loss $\mathcal{L}_{div}$ defined as follows: 
\begin{equation}
	\begin{aligned}
		\label{eq:diversity_loss}
		\mathcal{L}_{div} & = -\frac{1}{|R|}\sum\limits_{r=1}^{|R|}||\bm{\mu}_{r,\cdot}-\tilde{\bm{\mu}}_{r,\cdot}|| - \frac{1}{N}\sum\limits_{s=1}^{N}||\bm{\mu}_{\cdot,s}-\tilde{\bm{\mu}}_{\cdot,s}||,  \\
		\text{where }& \; \bm{\mu}_{r,\cdot} =  \frac{1}{N}\sum\limits_{s=1}^{N}\mathbf{a}_{r,s}(x),\;\; \bm{\mu}_{\cdot,s} =  \frac{1}{|R|}\sum\limits_{r=1}^{|R|}\mathbf{a}_{r,s}(x), 
	\end{aligned}
\end{equation}
where $\mathbf{a}_{r,s}(x)$ denotes the adaptiveness factor for the $s$-th sampled point on the $r$-th ray. 
$\tilde{\bm{\mu}}_{r,\cdot}$ and $\tilde{\bm{\mu}}_{\cdot,s}$ are randomly shuffled from $\{\bm{\mu}_{r,\cdot}\}_{r=1}^{|R|}$ and 
$\{\bm{\mu}_{\cdot,s}\}_{s=1}^{N}$ respectively, where the randomness in shuffling suppresses the trivial solution. 

The final target function of OPONeRF is the summation of photometric loss $\mathcal{L}_{pho}$ for volume rendering, reconstruction loss $\mathcal{L}_{rec}$, approximation loss $\mathcal{L}_{appr}$ (KL divergence) for point representation, and diversity loss $\mathcal{L}_{div}$ for parameters of renderer: 
\begin{equation}
	\label{eq:final_loss}
	\mathcal{L}_{opo} = \mathcal{L}_{pho} + \mathcal{L}_{appr} + \gamma\mathcal{L}_{rec} + 1e^{-5}\mathcal{L}_{div}. 
\end{equation}

\subsection{Implementation Details}
\label{sec:implementation_details}
In this subsection, we present the implementation details and the design choices in OPONeRF. 
We follow the MVSNet~\citep{yao2018mvsnet} architecture to build cost volumes and the overall scene representation. 
In each PCD, both $D^l_w(\cdot)$ and $D^l_m(\cdot)$ contain three consecutive fully-connected layers. 

\textbf{Full-sized Output of PCD: }
As the width of a target layer can be very large ($C^l_i,C^l_o>1000$), the computational cost can be expensive for the parameter candidate decoders if we obtain the full-sized output in Eqn.~(\ref{eq:soft_mask_generation}) at once. 
Therefore, we firstly generate a down-sized output of Eqn.~(\ref{eq:soft_mask_generation}) then apply linear layers to upscale the fan-in and fan-out dimensions to their raw resolution. 
Inspired by \citep{chen2022transformers}, we element-wisely multiply the upscaled output with another set of full-sized matrices that are randomly initialized and iteratively optimized. 
The similar technique is adopted in 3D object detection~\citep{zheng2022hyperdet3d}, where the additional full-sized matrices can be regarded as the scene-agnostic representation.

\textbf{Binary Quantization in $\bm{D^l_m}$: }
As the step function $S(x)$ for the quantization of $M^l$ is non-differentiable, we implement its gradient as a long-tailed estimator $H(x)$~\citep{liu2020dynamic}: 
\begin{equation}
	\label{eq:approximation}
	\frac{d}{dx}S(x) \approx H(x) =
	\begin{cases}
		2-4|x|,  & -0.4 \le x \le 0.4 \\
		0.4,     & 0.4 < |x| \le 1 \\
		0,       & otherwise 
	\end{cases}
\end{equation}

\textbf{Incorporating OPONeRF into Other Methods: }
The idea of OPONeRF can be flexibly incorporated into other popular methods. 
We set the target layers as all linear layers but the last regression heads. 
For example, to implement One-Point-One MVSNeRF (OPO-MVSNeRF), the target layers are all layers except for the radiance and density MLPs in its NeRF renderer. 
We follow Sec.~\ref{sec:adaptiveness_decoder} to correspondingly learn a PCD module that generates candidate parameters for the target layers, then regress per-point adaptiveness factors to personalize neural renderers for each sampled point. 
We validated the effectiveness of the spirit of OPONeRF via implementing such baseline variants and conducting benchmark experiments. 
Please refer to Sec.~\ref{sec:discussion_and_ablation} for details. 

\subsection{Discussions}
\label{sec:discussion}
In this subsection, we present a more in-depth analysis of OPONeRF. 
We also discuss the relation and differences between related works in a more detailed manner. 

\begin{figure}[bt!]
	\centering
	%\vspace{-2mm}
	\includegraphics[width=0.48\textwidth]{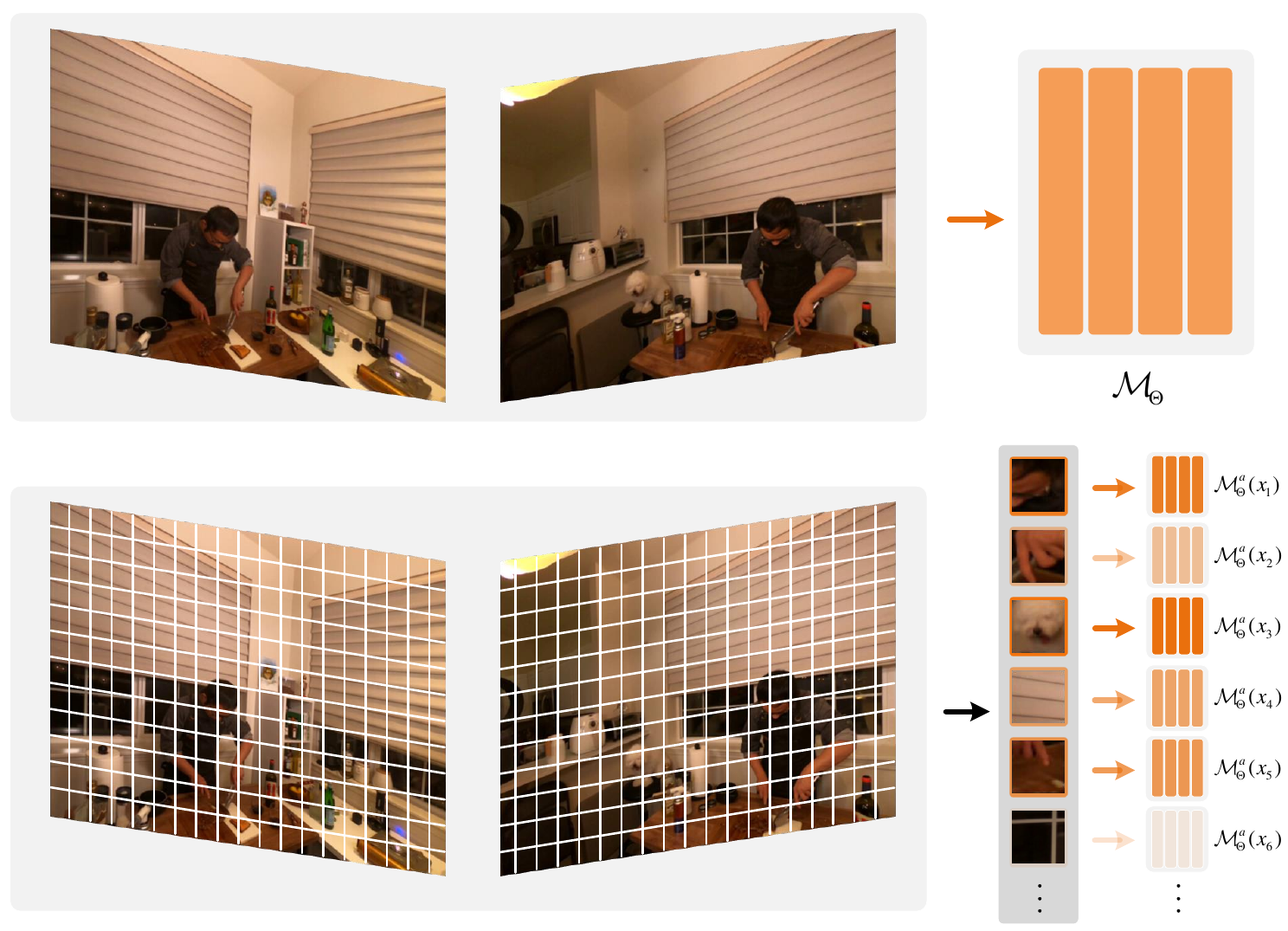}
	%\put(10,10){hello}
	\vspace{-6mm}
	\caption{
		Upper: 
		conditioning the parameters of the renderer $\mathcal{M}_\Theta$ on the global representation of the given scene. 
		Lower: 
		learning to personalize a unique neural renderer $\mathcal{M}_\Theta^a(\mathbf{x}_i)$ for each densely sampled coordinate $\mathbf{x}_i$. 
		In this figure, we use 2D image patches instead of 3D local coordinates to represent $\mathbf{x}_i$s for intuitive illustration. 
		(\textit{Best viewed in color.})
	}
	\label{fig:local_vs_global}
\end{figure}

\textbf{Awareness of Variation from a Single Timeframe: }
In previous benchmark setups~\citep{li2022neural,gao2021dynamic}, the training data usually contains multiple timeframes. 
However, in real-life scenario, it is common to model a fixed 3D scene and encounter perturbations such as unseen motions, illumination variations or data contaminations at test time. 
We therefore investigate training on a single frozen timeframe which is conceptually more challenging. 
Under this setting, if we ignore the adaptiveness factor $\mathbf{a}_x$ and directly set: 
\begin{equation}
	\label{eq:no_local}
	\mathbf{W}^l_x := \mathbf{W}^l_a, \forall x , 
\end{equation}
then the adaptiveness of neural renderer to the test-time scene variations is limited by the one-shot learning framework. 
This is because we simply condition the parameters of neural renderer ($\mathcal{M}_\Theta$) on the \textbf{global} representation of the given scene, as shown in the upper part in Fig.~\ref{fig:local_vs_global}. 
%(Sec.~\ref{sec:geometry_encoder}+Sec.~\ref{sec:adaptiveness_decoder}).
Therefore, changes can hardly be captured by training with a single and frozen timeframe. 
However, the incorporation of \textbf{local} adaptiveness factors converts the non-trivial one-shot learning into a more generalizable framework. 
As shown in the lower part of Fig.~\ref{fig:local_vs_global}, OPONeRF learns to personalize a unique renderer for every sampled $x$.   %(Sec.~\ref{sec:geometry_encoder}+Sec.~\ref{sec:adaptiveness_decoder}+Sec.~\ref{sec:low_dimensional_factor}) further decides how to personalize candidate layers for queried positions. 
By covering all the densely sampled 3D positions, OPONeRF hence learns the respond to the abundant local variations even from a single frozen timeframe, broadening the concept of seen variations from scene-level adaptation to point-level generalization. 
Since our experimental setting focuses on local scene changes, we assume that the awareness of rich training-time local variation is beneficial to the adaptiveness and robustness to those test-time changes. 

\textbf{Comparison with NeRF-VAE: }
Variational inference is also employed in NeRF-VAE~\citep{kosiorek2021nerf}. 
Equipped with the generative model, NeRF-VAE is able to sample novel scenes and viewpoints. 
However, in contrast to the local subdivisions of a single training scene in OPONeRF, NeRF-VAE relies on abundant scene-level variations (100k or 200k scenes~\citep{kosiorek2021nerf}) to build a sizable sampling space. 
Moreover, the expressiveness of NeRF-VAE is determined by the dimension of the sampled latent code. 
In OPONeRF, the flexibility is reflected by the size of parameter matrices which are on a larger scale, hence enabled with larger information capacity. 
Besides, NeRF-VAE conditions a fixed renderer on a single latent variable similar to the generalization-based methods~\citep{trevithick2021grf,chen2021mvsnerf,liu2022neuray}. 
OPONeRF instead explicitly decomposes the point representation into deterministic invariance and unexpected variance which dynamically personalizes a point-wise renderer.

\textbf{Comparison with KiloNeRF: }
KiloNeRF~\citep{reiser2021kilonerf} also uses a divide-and-conquer strategy where plenty of NeRFs take charge of subdivided 3D blocks. 
The difference is that OPONeRF adaptively responds to local feature variations while the independent MLPs in KiloNeRF are passively assigned to each block beforehand~\citep{zhu2023pyramid}. 
Moreover, the personalized renderer in OPONeRF maintains the network capacity to take the uncertainty into account, while the MLPs in KiloNeRF are in tiny scales for acceleration. 
As for the functionality, KiloNeRF overfits to a single scene while OPONeRF can generalize to unseen changed scenes. 

\begin{figure}[tb]
	\centering
	\includegraphics[width=0.48\textwidth]{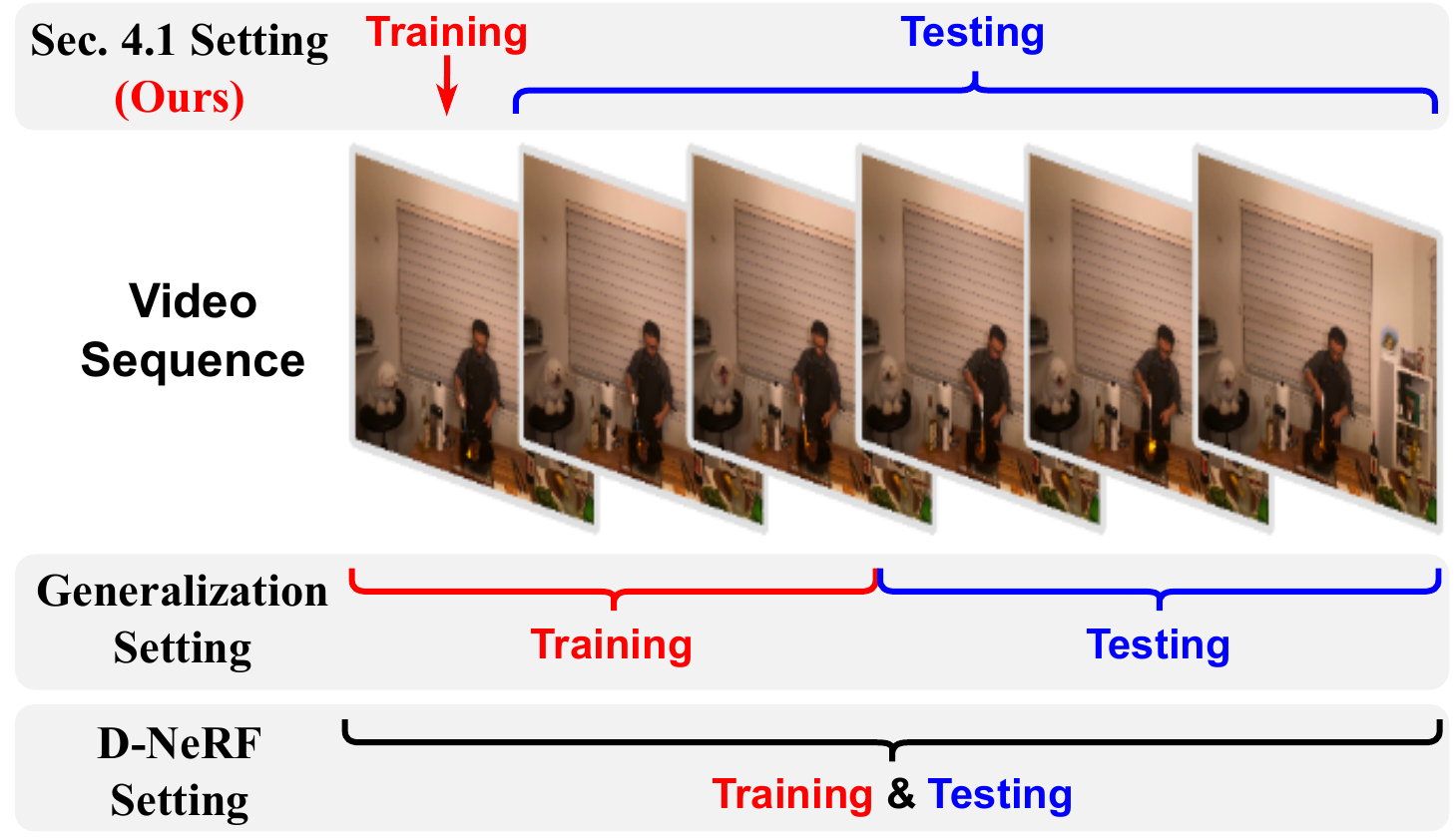}
	\vspace{-6mm}
	\caption{
		Layout of different experimental benchmarks. For easy illustration, we only show a single-view video sequence here instead of multi-view used in our benchmark. 
	} \label{fig:setting_comparison}
\end{figure}

\section{Experiment}
\begin{table*}[ht]
	\caption{Quantitative comparison with baseline methods on the proposed Cooking-Perturbation benchmark. 
		The grayed rows indicate the quantitatively and visually defective or even failed results on our benchmark. 
	}
	\vspace{-2mm}
	\centering
	\resizebox{.98\textwidth}{!}{
		\begin{tabular}{lcccccccccccc}
			\toprule
			& \multicolumn{3}{c}{Cut Roasted Beef} & \multicolumn{3}{c}{Cook Spinach} & \multicolumn{3}{c}{Flame Steak} & \multicolumn{3}{c}{Sear Steak} \\
			Method & PSNR$\uparrow$ & SSIM$\uparrow$ & LPIPS$\downarrow$ & PSNR$\uparrow$ & SSIM$\uparrow$ & LPIPS$\downarrow$ & PSNR$\uparrow$ & SSIM$\uparrow$ & LPIPS$\downarrow$
			& PSNR$\uparrow$ & SSIM$\uparrow$ & LPIPS$\downarrow$\\
			\midrule
			\rowcolor{Gray}
			GeoNeRF~\citep{johari2022geonerf} 
			& 18.95 & 0.538 & 0.253 & 17.91 & 0.513 & 0.272 & 19.05 & 0.517 & 0.277 & 17.05 & 0.408 & 0.490\\
			\rowcolor{Gray}
			GNT~\citep{wang2022attention}
			& 20.25 & 0.487 & 0.730 & 18.38 & 0.501 & 0.700 & 18.35 & 0.493 & 0.718 & 18.47 & 0.492 & 0.714 \\
			MVSNeRF~\citep{chen2021mvsnerf}
			& 23.31 & 0.546 & 0.278 & 22.97 & 0.574 & 0.273 & 23.87 & 0.586 & 0.256 & 23.87 & 0.586 & 0.256\\
			IBRNet~\citep{wang2021ibrnet} & 26.39 & 0.803 & 0.238 & 24.27 & 0.709 & 0.182 & 25.71 & 0.762 & 0.225 & 23.86 & 0.747 & 0.276 \\
			NeuRay~\citep{liu2022neuray}
			& 28.52 & 0.896 & 0.181 & 29.38 & 0.912 & 0.161 & 30.33 & 0.926 & 0.141 & 30.95 & \textbf{0.937} & 0.122
			\\ \midrule

			OPONeRF & \textbf{29.68} & \textbf{0.916} & \textbf{0.154} &  \textbf{31.81} & \textbf{0.935} & \textbf{0.121} & \textbf{31.95} & \textbf{0.941} & \textbf{0.116}  & \textbf{31.95} & \textbf{0.937} & \textbf{0.119} \\
			\bottomrule
		\end{tabular}
	}
	\label{tab:results_video_dataset}
\end{table*}

\begin{table*}[ht]
	\caption{Quantitative comparison with baseline methods on the proposed MeetRoom-Perturbation benchmark. 
		The grayed rows indicate that GeoNeRF~\citep{johari2022geonerf} and GNT~\citep{wang2022attention} visually \textbf{fail} on our benchmark.  
	}
	\vspace{-2pt}
	\centering
	\resizebox{.86\textwidth}{!}{
		\begin{tabular}{lccccccccc}
			\toprule
			& \multicolumn{3}{c}{Discussion} & \multicolumn{3}{c}{Trimming} & \multicolumn{3}{c}{VR Headset} \\
			Method & PSNR$\uparrow$ & SSIM$\uparrow$ & LPIPS$\downarrow$ & PSNR$\uparrow$ & SSIM$\uparrow$ & LPIPS$\downarrow$ & PSNR$\uparrow$ & SSIM$\uparrow$ & LPIPS$\downarrow$\\
			\midrule
	
			\rowcolor{Gray}
			GeoNeRF~\citep{johari2022geonerf} 
			& 6.94 & 0.001 & 0.996 & 9.52 & 0.007 & 1.011 & 6.95 & 0.003 & 1.007 
			\\
			\rowcolor{Gray}
			GNT~\citep{wang2022attention}
			& 13.84 & 0.569 & 0.575 & 11.20 & 0.483 & 0.625 & 11.51 & 0.452 & 0.618\\
			\rowcolor{Gray}
			MVSNeRF~\citep{chen2021mvsnerf}
			& 11.29 & 0.445 & 0.437 & 12.26 & 0.439 & 0.456 & 12.39 & 0.443 & 0.438\\
			NeuRay~\citep{liu2022neuray}
			& 27.79 & 0.932 & 0.149 & 29.88 & 0.942 & 0.141 & 28.61 & 0.937 & 0.153
			\\ \midrule
			
			OPONeRF  
			& \textbf{28.03} & \textbf{0.937} & \textbf{0.144} & \textbf{30.26} & \textbf{0.948} & \textbf{0.132} &   \textbf{30.13} & \textbf{0.944} & \textbf{0.144}\\
			\bottomrule
		\end{tabular}
	}
	\label{tab:results_meetroom_dataset}
\end{table*}

In this section, we firstly conducted extensive experiments to evaluate our OPONeRF on our constructed benchmarks for test-time perturbation, including foreground movement, illumination changes and multi-model noises. 
Moreover, we have validated the effectiveness of OPONeRF by evaluating on the conventional generalization-based benchmarks. 
We finally analyze the design choice and effectiveness of OPONeRF through ablation studies and qualitative illustration. 
We have also tested the general applicability of our idea by incorporating OPONeRF into other popular baselines to get consistent improvements. 

We evaluated the results using PSNR, SSIM~\citep{hore2010image} and LPIPS~\citep{zhang2018unreasonable} as metrics. 
``OPO-'' denotes the baseline method incorporated with the proposed OPONeRF. 
Numbers in {\color{red} red} and {\color{DarkGreen} green} indicate performance improvement and degradation respectively. 
Training and inference of all experiment trials were done on a single Nvidia RTX3090 24GB GPU. 
\subsection{Experiments on Test-time Foreground Movements}
We firstly construct benchmarks with foreground motions which are inevitable in physical world. 
\subsubsection{Benchmark Construction}
\label{sec:datasets_and_settings_realistic}
We firstly constructed two benchmarks re-organized from the real-world Plenoptic Cooking Video~\citep{li2022neural} and MeetRoom~\citep{li2022streaming} datasets, named Cooking-Perturbation and MeetRoom-Perturbation respectively. 
We then utilized the \textbf{Kubric} dataset generator engine~\citep{greff2022kubric} to simulate the object adding or removal, named Kubric-Perturbation respectively. 

\textbf{Cooking-Perturbation benchmark: }
The raw Plenoptic Cooking Video~\citep{li2022neural} dataset contains multi-view indoor videos captured by 21 GoPro cameras. 
The multi-view cameras capture a standing person conducting a certain task. 
To exclude the light interference, we selected the nighttime tasks of ``Cook Spinach'', ``Cut Roasted Beef'', ``Flame Steak'' and ``Sear Steak'' to construct our benchmark. 
We keep the image resolution as the original 2704x2028. 

\textbf{MeetRoom-Perturbation benchmark: }
The raw MeetRoom~\citep{li2022streaming} dataset is recorded by 13 Azure Kinect cameras. 
All task scenes (``Discussion'', ''VR Headset'' and ``Trimming'') are adopted in our benchmark. 
The movements in each scene are in larger magnitude and more complicated compared to \citep{li2022neural}, especially for the task ``Discussion'' where multiple persons are interacting. 
We keep the image resolution as the original 1280x720. 

\textbf{Settings: }
For all tasks in two benchmarks, we evenly sampled 21 frames along the timeline for each camera and used the first timeframe for training and validation. 
All views in the rest 20 timeframes are adopted as test set. 
Note that the evaluation on the validation set  equals the classic \textbf{per-scene optimization} benchmark as in~\citep{mildenhall2020nerf}. 
We briefly illustrate the experimental settings in Fig.~\ref{fig:setting_comparison}, where only one viewpoint is shown for simplicity. 

We used 5 evenly-indexed camera views in the first timeframe for training. 
The compared beselines mainly consist of IBRNet~\citep{wang2021ibrnet}, MVSNeRF~\citep{chen2021mvsnerf},  NeuRay~\citep{liu2022neuray}, GeoNeRF~\citep{johari2022geonerf} and GNT~\citep{wang2022attention}. 

\begin{figure}[bt!]
	\centering
	\includegraphics[width=0.48\textwidth]{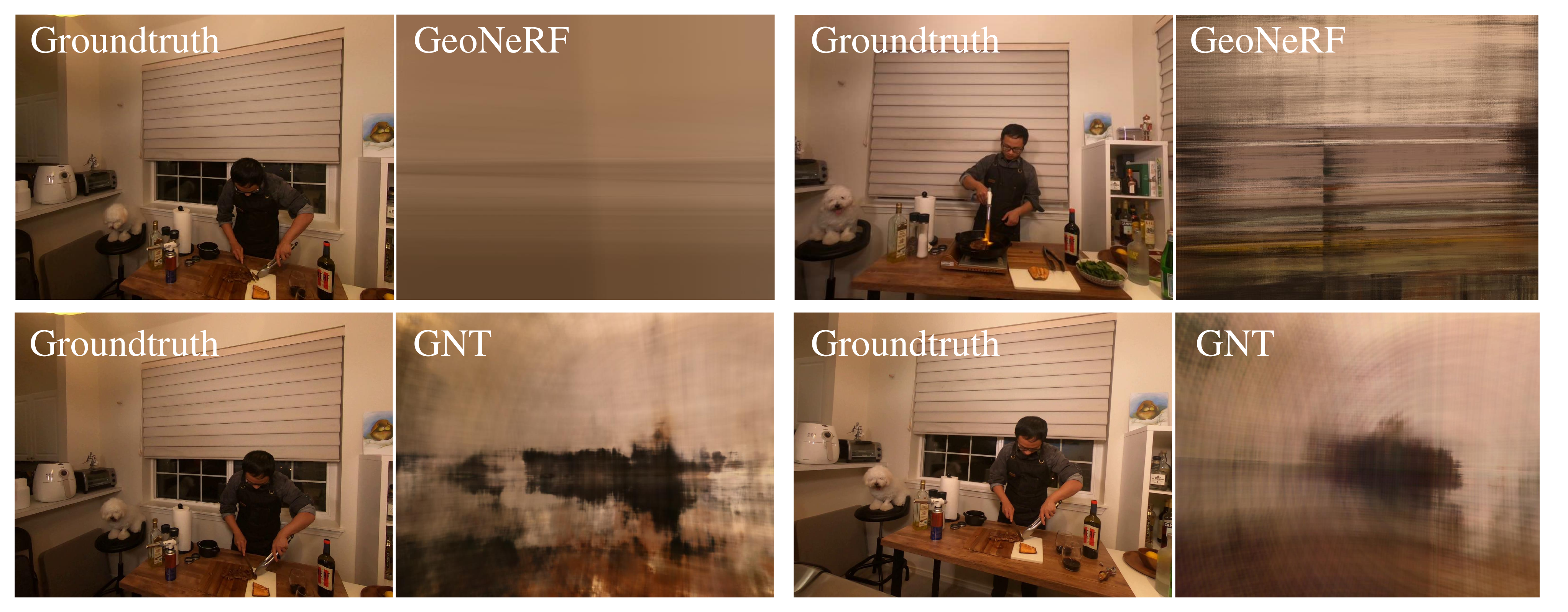}
	\vspace{-7mm}
	\caption{
		Advanced generalization-based methods such as GeoNeRF~\citep{johari2022geonerf} and GNT~\citep{wang2022attention} fail when confronted with test-time foreground movements. 
	} \label{fig:failed_example}
\end{figure}

\subsubsection{Experimental Results and Analysis }
%The compared methods mainly consist of MVSNeRF~\citep{chen2021mvsnerf}, IBRNet~\citep{wang2021ibrnet}, NeuRay~\citep{liu2022neuray}, GeoNeRF~\citep{johari2022geonerf} and GNT~\citep{wang2022attention}. 
The quantitative results on Cooking-Perturbation and MeetRoom-Perturbation are shown in Table~\ref{tab:results_video_dataset} and Table~\ref{tab:results_meetroom_dataset} respectively. 
%We split the evaluation of rendered results into training views and novel views. 
The observation is 2-fold. 
Firstly, while GeoNeRF and GNT learn stronger scene priors via advanced architectures, they demonstrate much inferior rendering performance when tackling test-time foreground movements as shown qualitative in Fig.~\ref{fig:failed_example}. 
Also, as shown in Fig.~\ref{fig:qualitative_main}, GeoNeRF produces extremely blurred results on Cooking-Perturbation and even renders all-black images on MeetRoom-Perturbation. 
This indicates that the stronger generalization ability cannot guarantee the robustness to small but unpredictable test-time scene changes. 
It also demonstrates that our benchmark, widely existing in daily life, is much more challenging than the conventional generalization-based settings. 
In contrast, methods with fewer scene priors and less sophisticated architectures output visually better results on Cooking-Perturbation, such as MVSNeRF and NeuRay. 
Note that MVSNeRF still struggles on MeetRoom-Perturbation. 
Secondly, quantitatve results show that our OPONeRF consistently achieves state-of-the-art results on all 3 metrics, which validates the effectiveness of the proposed method. 
With sufficient training samples for the mapping from point to parameters of neural renderer, OPONeRF is able to adapt to unseen changed foreground and preserve the background details with only a single training timeframe. 
Notably, on ``Cook Spinache'', ``Flame Steak'' and ``VR Headset'' tasks, OPONeRF outperforms the second-best NeuRay by a sizable margin ($>$1.5 PSNR). 

\begin{table*}[ht]
	\centering
	\caption{Quantitative comparison with baseline methods by pretraining on the task of ``Cook Spinach'' and evaluating on the other tasks. 
	}
	\vspace{-2pt}
	\resizebox{.87\textwidth}{!}{
		\begin{tabular}{lcccHHHcccccc}
			\toprule
			& \multicolumn{3}{c}{Cut Roasted Beef} & \multicolumn{3}{H}{Cook Spinach} & \multicolumn{3}{c}{Flame Steak} & \multicolumn{3}{c}{Sear Steak} \\
			Method & PSNR$\uparrow$ & SSIM$\uparrow$ & LPIPS$\downarrow$ & PSNR$\uparrow$ & SSIM$\uparrow$ & LPIPS$\downarrow$ & PSNR$\uparrow$ & SSIM$\uparrow$ & LPIPS$\downarrow$
			& PSNR$\uparrow$ & SSIM$\uparrow$ & LPIPS$\downarrow$\\
			\midrule
			\rowcolor{Gray}
			GNT~\citep{wang2022attention} & 17.94  & 0.704 & 0.496 & 16.45 & 0.912 & 0.161 & 16.45 & 0.692 & 0.506 & 16.43 & 0.689 & 0.506\\
			\rowcolor{Gray}
			GeoNeRF~\citep{johari2022geonerf} & 17.04 & 0.544 & 0.470 & 29.38 & 0.912 & 0.161 & 18.75 & 0.546 & 0.264 & 16.38 & 0.517 & 0.501\\
			\rowcolor{Gray}
			MVSNeRF~\citep{chen2021mvsnerf}
			& 20.23 & 0.655 & 0.266 & 29.38 & 0.912 & 0.161 & 20.62 & 0.671 & 0.293 & 22.22 & 0.701 & 0.174\\
			% & LLFF~\citep{mildenhall2019local}
			% & 24.77 & 0.911 & 0.114 &    -  &     - &     - & 24.41 & 0.805 & 0.211 \\
			NeuRay~\citep{liu2022neuray}
			& 28.86 & 0.907 & 0.165 & 29.38 & 0.912 & 0.161 & 30.03 & 0.921 & 0.155 & 30.08 & 0.921 & 0.155 \\
			%OPO-NeuRay& 31.59 & 0.932 & 0.123 & 31.63 & 0.934 & 0.123 & 32.06 & 0.937 & 0.121 & 32.11 & 0.937 & 0.121\\
			\midrule
			OPONeRF & \textbf{31.76} & \textbf{0.934} & \textbf{0.120}
			& \textbf{31.63} & \textbf{0.934} & \textbf{0.123} & \textbf{32.36} & \textbf{0.941} & \textbf{0.116} & \textbf{32.37} & \textbf{0.940} & \textbf{0.116}\\
			\bottomrule
		\end{tabular}
	}
	\label{tab:results_cross_task}
\end{table*}

To simulate more sophisticated foreground movements in the same environment, we also conducted \textbf{cross-task} evaluation by training on the ``Cook Spinach'' task and evaluating on the other 3 tasks in the Cooking-Perturbation benchmark. 
Results in Table~\ref{tab:results_cross_task} and Fig.~\ref{fig:qualitative_main} show that OPONeRF demonstrates substantially superior robustness to more complex test-time motions. 
It is worth noting that while intuitively cross-task rendering is more challenging than within-task, both NeuRay and OPONeRF achieve better results on the cross-task inference compared with Table~\ref{tab:results_video_dataset}. 
Given the fact that both training and testing timeframes share the same background and that the test-time motions are unpredictable, we postulate that little difference lies between the settings of cross-task and within-task inference. 

As indicated by D-NeRF setting~\citep{pumarola2021d} in Fig.~\ref{fig:setting_comparison}, scene variations can also be modeled by the methods designed for the spatio-temporal reconstruction benchmarks~\citep{pumarola2021d,tian2023mononerf,gao2021dynamic}. 
However, they are built on the premise of multiple timeframes at the training phase. 
In our settings, the deformation mapping $(\mathbf{x}, \mathbf{d}, t)\rightarrow (\mathbf{c},\sigma)$ in D-NeRF downgrades to $(\mathbf{x}, \mathbf{d})\rightarrow (\mathbf{c},\sigma)$ in the absence of temporal variation. 
Therefore, the temporal modeling in D-NeRF is infeasible in our constructed benchmark, which we quantitatively show in Sec.~\ref{sec:discussion_and_ablation}.

\begin{table}
	\centering
	\caption{Quantitative comparison with baseline methods on the robustness to synthetic test-time illumination changes. 
	}
	\vspace{-2pt}
	\resizebox{.48\textwidth}{!}{
		\begin{tabular}{lcccccc}
			\toprule
			&\multicolumn{3}{c}{Scene set $W$} &\multicolumn{3}{c}{Scene set $B$}\\
			Method & PSNR$\uparrow$ & SSIM$\uparrow$ & LPIPS$\downarrow$ & PSNR$\uparrow$ & SSIM$\uparrow$ & LPIPS$\downarrow$ \\ \midrule
			IBRNet~\citep{wang2021ibrnet} & 25.15 & 0.795 & 0.113 & 26.88 & 0.813 & 0.089\\
			MVSNeRF~\citep{chen2021mvsnerf} & 31.15 & 0.902 & 0.050 & 31.85 & 0.947 & 0.070 \\
			NeuRay~\citep{liu2022neuray} & 33.52 & 0.975 & 0.049 & 34.54 & 0.973 & \textbf{0.054} \\ \midrule
			OPONeRF & \textbf{35.23}& \textbf{0.978} & \textbf{0.035} & \textbf{34.99} & \textbf{0.975} & 0.067  \\ \bottomrule
		\end{tabular}
	}
	
	\label{tab:quantitative_results_clevr}
\end{table}
\begin{figure*}[tb]
	\centering
	\includegraphics[width=0.98\textwidth]{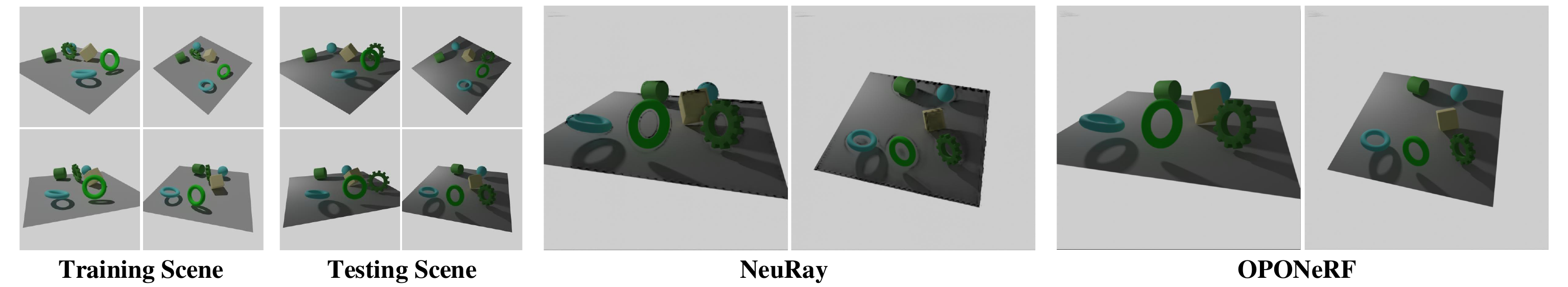}
	\vspace{-4mm}
	\caption{
		A snippet of the benchmark on test-time illumination changes using synthetic dataset engine Kubric~\citep{greff2022kubric}, and qualitative comparison of rendering results. 
		Scene changes including object movement and illumination variation can be observed by comparing training and testing scenes. 
		During inference, NeuRay produces shading artifacts around both static and moving objects while our OPONeRF better preserves the lighting conditions. 
	} \label{fig:clevr_qualititative}
\end{figure*}
	
	\subsection{Experiments on Test-time Illumination Changes}
	Then we simulate the test-time illumination changes which broadly exist due to the training-testing temporal difference. 
	\label{sec:datasets_and_settings_synthetic}
	\subsubsection{Benchmark Construction}
	We construct a database feasible for manual illumination variation. 
	Equipped with the \textbf{Kubric} dataset generator engine~\citep{greff2022kubric}, we randomly initialized 2 scenes with white and black background respectively, named scene $W$ and $B$. 
	Then we sequentially rendered 21 images ($\mathcal{I}_t^W = \{I^W_{i,t}\}^{20}_{i=0}$ and $\mathcal{I}^B_t = \{I^B_{i,t}\}^{20}_{i=0}$, $t=0$) from 2 scenes along a pre-set virtual viewpoint trajectory while keeping the scenes static. 
	%The depth values off the plane and in the background were clipped at 20m. 
	After that, we randomly changed both \textbf{overall} illumination in the scene and location of an object for 10 times, getting another 10 changed scenes from $W$ and $B$ respectively, i.e. $\{\mathcal{I}_t^W\}^{10}_{t=1}$ and $\{\mathcal{I}^B_t\}^{10}_{t=1}$. 
	The location and viewpoint of the cameras keep unchanged before and after scene perturbations. 
	We set the resolution of rendered images as $800\times800$. 
	
	\textbf{Settings: }
	Similar to Sec.~\ref{sec:datasets_and_settings_realistic}, we used the initial scene ($t=0$) and its changed versions ($t>0$) as the training\&validation scene and test scenes respectively. 
	We conducted benchmark experiments on IBRNet~\citep{wang2021ibrnet}, MVSNeRF~\citep{chen2021mvsnerf}, NeuRay~\citep{liu2022neuray} and our OPONeRF. 
	%We trained the model with 60k iterations and set the initial learning rate as 4e-5. 
	%The number of training views was set as 10. 

	\subsubsection{Experimental Results and Analysis: }
	The quantitative and qualitative results are summarized in Table~\ref{tab:quantitative_results_clevr} and Fig.~\ref{fig:clevr_qualititative} respectively. 
	It can be seen that the overall performance on the synthetic dataset is much better than real-world benchmarks numerically. 
	We infer it is due to the relatively low geometric complexity, so that the baselines with generalization power can well handle the test-time object movements. 
	However, we observe abnormal shadows around both the static and moving objects in the rendered results of NeuRay, which indicates the confusion of illumination conditions at test-time while OPONeRF more accurately preserves the lighting conditions through illustrative comparisons. 
	Quantitatively, in both backgrounds, OPONeRF shows substantially better numerical results in 3 evaluation metrics, regardless of the background color. 
	\begin{figure}[bt!]
		\centering
		\includegraphics[width=0.49\textwidth]{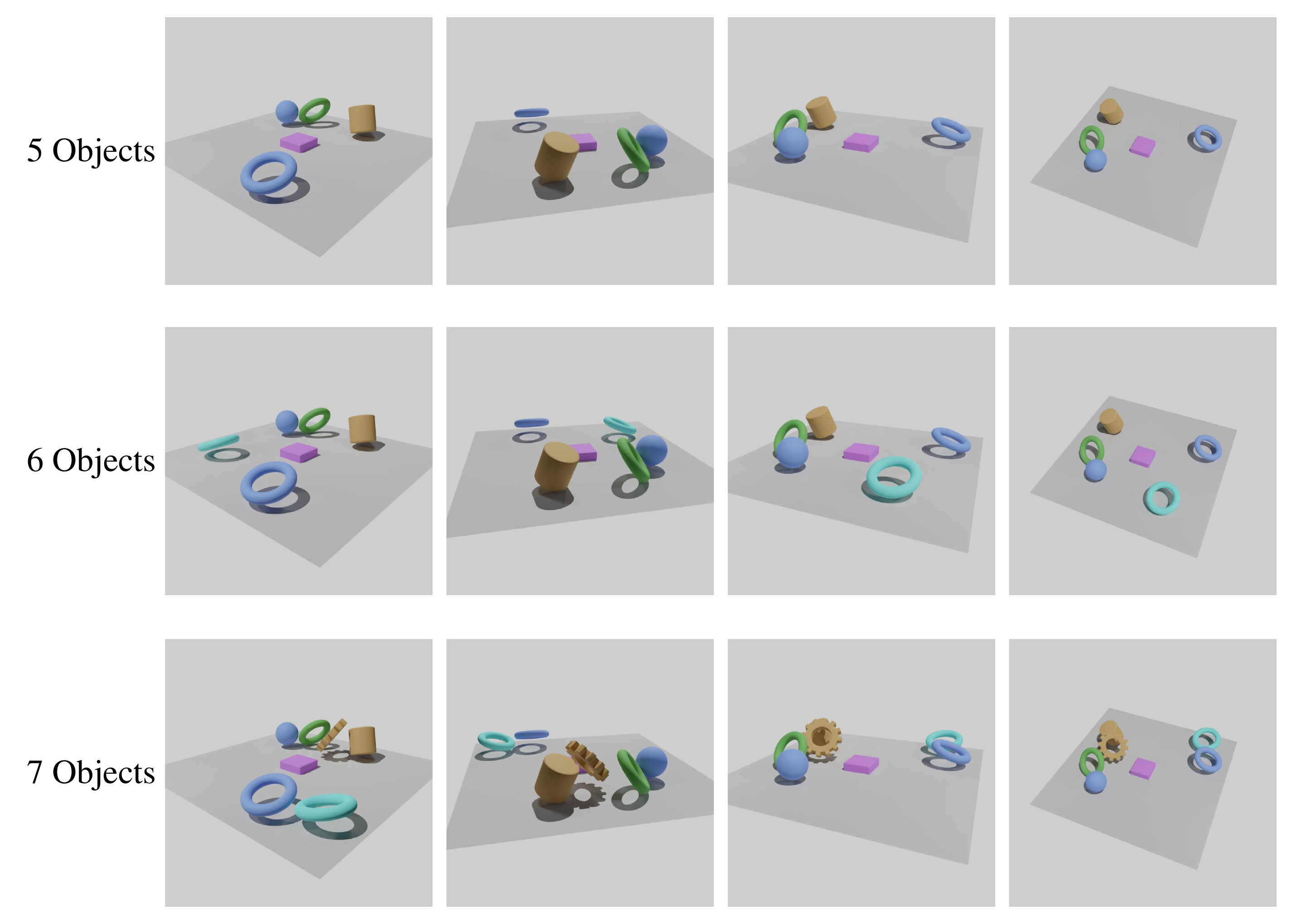}
		\vspace{-6mm}
		\caption{
			%	Qualitative comparison between state-of-the-art methods~\citep{wang2021ibrnet,liu2022neuray} and OPONeRF on LLFF. 
			%Qualitative comparison on LLFF. 
			Four of 21 groundtruth views of the benchmark on modifying object existence(s) using the Kubric engine~\citep{greff2022kubric}. 
			%	(\textit{Best viewed when zoomed in.})
		} \label{fig:modify_existence_gt}
	\end{figure}
	\begin{table}[bt!]
		\centering
		\setlength{\tabcolsep}{5.3pt}
		\caption{Quantitative results of the setting of modifying the object existence(s). The results are evaluated by PSNR$\uparrow$. }
		\vspace{-2mm}
		\begin{tabular}{cc|ccc}
			\toprule
			\multicolumn{2}{c|}{\multirow{2}{*}{OPONeRF/NeuRay}} & \multicolumn{3}{c}{Trained on} \\ 
			& & 5 Objects & 6 Objects & 7 Objects \\ \midrule
			\multirow{3}*{Tested on} & 5 Objects & \textbf{38.87}/36.36 & \textbf{39.27}/36.14 & \textbf{37.78}/32.27\\
			& 6 Objects & \textbf{38.11}/35.85 & \textbf{38.43}/35.55 & \textbf{37.42}/32.37\\
			& 7 Objects & \textbf{37.35}/35.32 & \textbf{37.71}/35.09 & \textbf{36.97}/32.06\\ \bottomrule
		\end{tabular}
		\label{tab:modify_existence}
	\end{table}
	\begin{figure*}
		\begin{center}
			\begin{minipage}{0.63\linewidth}
				\includegraphics[width=\linewidth]{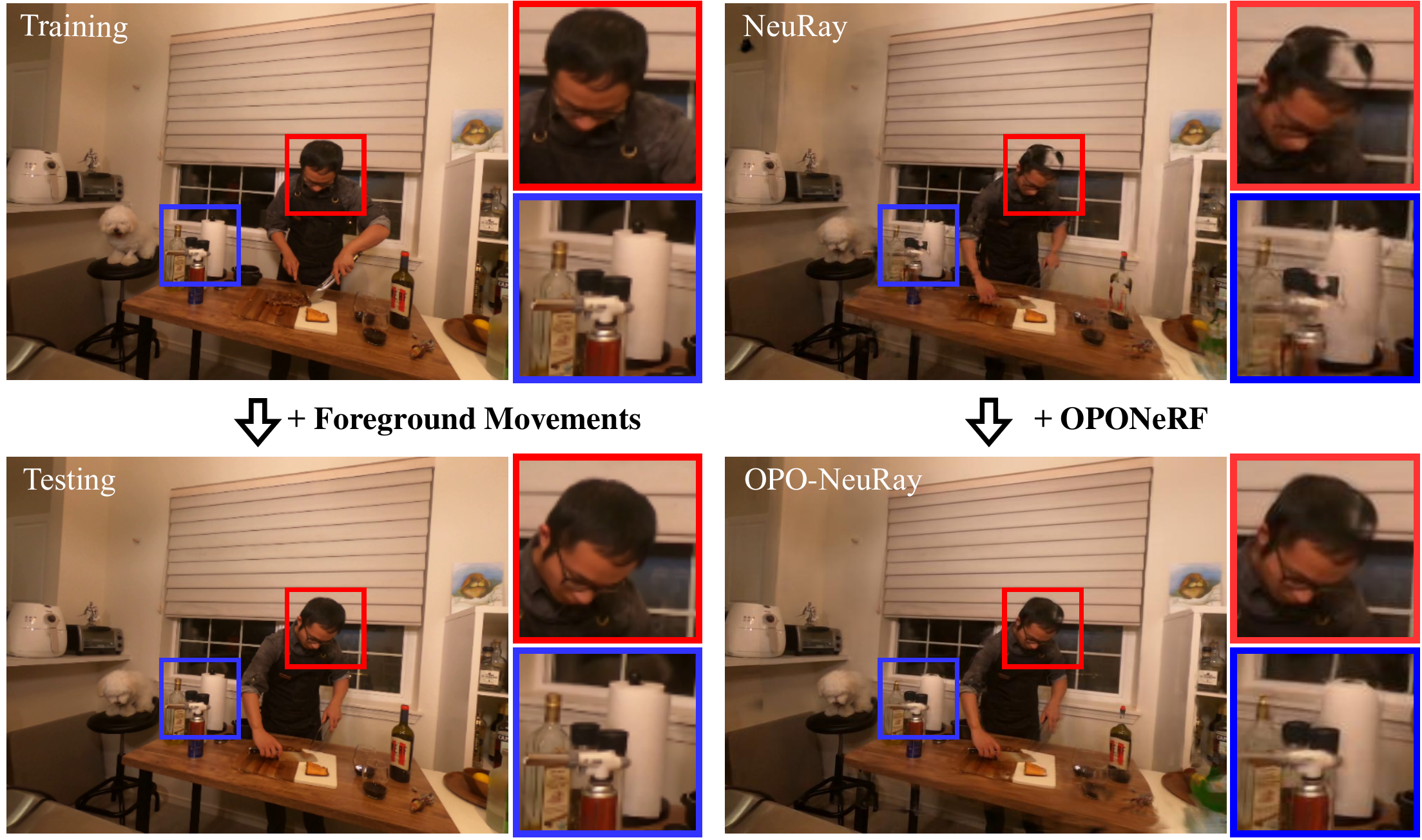}
				\vspace{-7mm}
				\captionof*{figure}{(a) Robustness to foreground movements. }
				\label{fig:teaser_left}
			\end{minipage}
			\begin{minipage}{0.348\linewidth}
				\includegraphics[width=\linewidth]{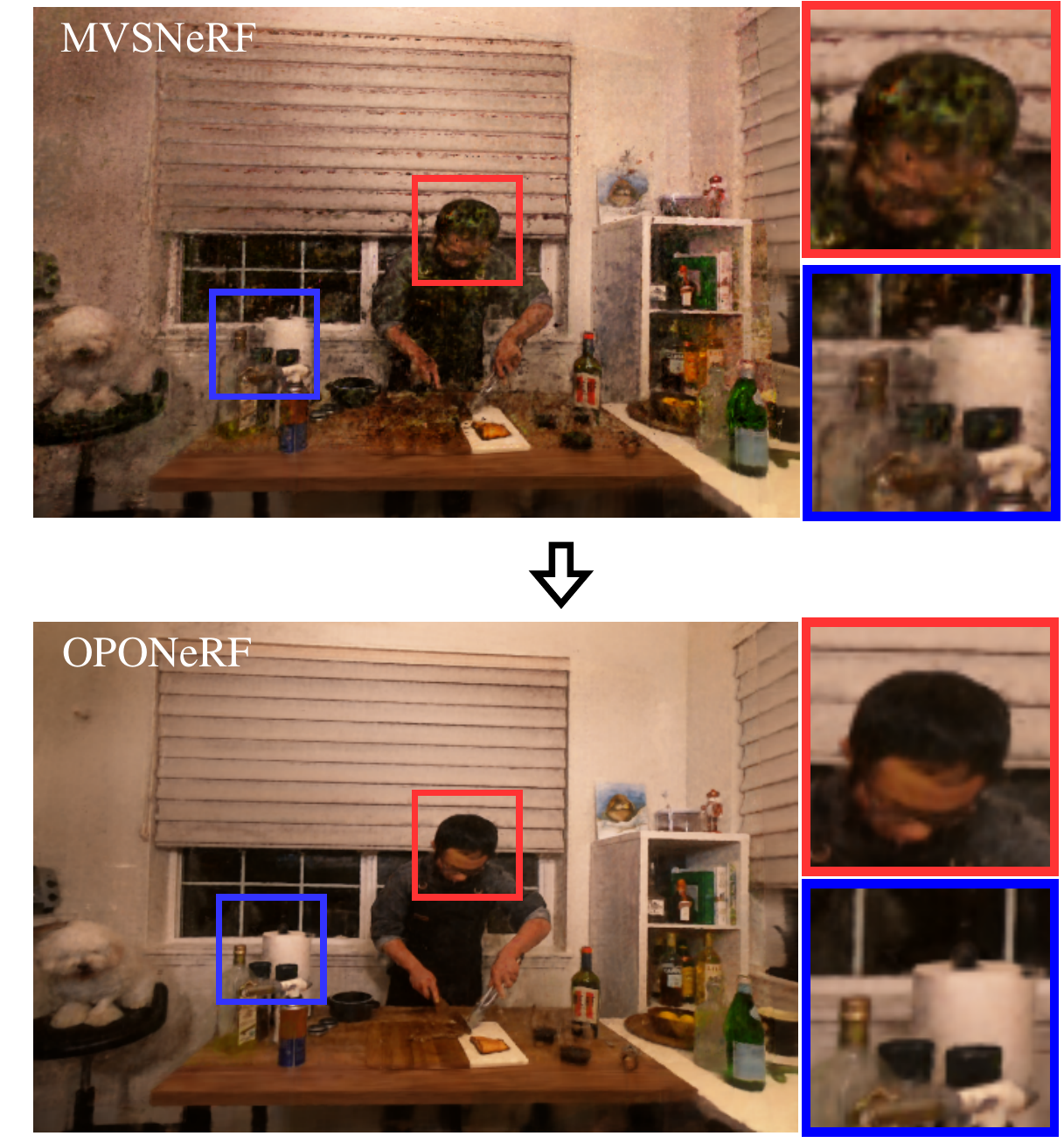}
				\vspace{-7mm}
				\captionof*{figure}{(b) Robustness to data contaminations. }
				\label{fig:teaser_right}
			\end{minipage}
			\vspace{-2mm}
			\captionof{figure}{
				(a) Upper left: Raw scene for training. 
				(a) Bottom left: Perturbed scene for testing. By comparing training and testing scenes, foreground movements can be observed while the background is static. 
				(a) Right: Comparison between NeuRay~\citep{liu2022neuray} and OPONeRF on the robustness to scene changes. 
				NeuRay performs poorer at both foreground and background rendering under test-time scene changes, while OPO-NeuRay demonstrates its robustness in those areas. 
				(b): Comparison between MVSNeRF~\citep{chen2021mvsnerf} and OPONeRF on the robustness to test-time noises. 
				After imposing random noises at test time, MVSNeRF produces polluted results while OPONeRF keeps off most of the noises. 
				%(Some of the compared areas are highlighted with colored boxes. \textit{Best viewed when zoomed in.})
				Some of compared areas are highlighted with colored boxes. 
			}
			\label{fig:teaser}
			%\vspace{0.11cm}
		\end{center}
	\end{figure*}
		
	\subsection{Experiments on Test-time Object Existence Changes}
	We have also experimented on the robustness to test-time object existance changes via synthesizing object insertion or removal. 
	\subsubsection{Benchmark Construction}
	We used the Kubric engine~\citep{greff2022kubric} and 
	followed the settings in Sec.~\ref{sec:datasets_and_settings_synthetic}, except that we only modified the object existence(s). 
	We constructed three synthetic scenes with 5,6 and 7 objects respectively. 
	21 viewpoints are captured in each scene, some of which are shown in Figure~\ref{fig:modify_existence_gt}.

	\subsubsection{Experimental Results and Analysis}
	The quantitative results of cross-scene evaluations on the robustness to test-time object existence changes are summarized in Table~\ref{tab:modify_existence}. 
	We can see that compared with NeuRay~\citep{liu2022neuray}, our OPONeRF demonstrates much more robustness, regardless of object removal or insertion. 
	Moreover, as suggested by the diagonal cells in Table~\ref{tab:modify_existence}, OPONeRF precedes the compared method when the training and testing scenes are the same, which demonstrates the effectiveness of OPONeRF on the setting of per-scene optimization.

	\subsection{Experiments on Test-time Noises}
	We finally construct the benchmark with test-time data contamination by simulating the noises during data capturing or restoration. 

	\subsubsection{Benchmark Construction}
	We constructed a benchmark of test-time noises built upon Cooking-Perturbation, where we added online gaussian noises to the input support images or intermediate feature representation ($\mathbf{F}$) during inference. 
	Specifically, we jittered the RGB values or geometric feature $\mathbf{F}$ by adding gaussian noises with the magnitude of 1.5 and 0.3 respectively. 
	The inference and evaluation were on the validation set of ``Cut Roasted Beef'' scene, using the same pre-trained model as in the Cooking-Perturbation benchmark. 
	We compare OPONeRF with baselines including MVSNeRF~\citep{chen2021mvsnerf} and NeuRay~\citep{liu2022neuray}. 
	%Please refer to subsection~ for the details of OPO-MVSNeRF. 
	\begin{table}[bt!]
	\centering
	\caption{Quantitative comparison with baseline methods on the ``Cut Roasted Beef'' scene with test-time noises. The experiments were on the \textbf{validation} set, i.e. the per-scene optimization setting. The percentage of performance degradation is also shown. 
	}
	\vspace{-3pt}
	\resizebox{.47\textwidth}{!}{
		\begin{tabular}{cllHH}
			\toprule
			Settings & Method & PSNR$\uparrow$ & SSIM$\uparrow$ & LPIPS$\downarrow$\\
			\midrule
			% \multirow{2}{*}{Original}
			\multirow{3}{*}{Noise Free} & MVSNeRF~\citep{chen2021mvsnerf} & 27.44 &  0.881 & 0.075  \\
			%& OPO-NeuRay & 27.58 &  &  \\
			%& OPO-MVSNeRF & 27.91 & 0.885 & 0.070 \\ 
			& NeuRay~\citep{liu2022neuray} & 27.95 & 0.888 & \\
			& OPONeRF 
			& \textbf{28.50} & \textbf{0.902} & \textbf{0.158}\\ \midrule
			\multirow{3}{*}{Noise on Features}
			& MVSNeRF~\citep{chen2021mvsnerf}
			% & 24.45 & 0.758 & 0.146  \\
			& 22.36 {\color{DarkGreen} (-18.51\%)} & 0.623  & 0.270  \\
			%& OPO-NeuRay & 27.53 {\color{DarkGreen} (\textbf{-0.18\%})} & &  \\
			%& OPO-MVSNeRF & 26.66 {\color{DarkGreen} (-4.48\%)} & 0.833 & 0.098 \\
			& NeuRay~\citep{liu2022neuray} & 27.71 {\color{DarkGreen} (-0.86\%)} & 0.884 & \\
			& OPONeRF 
			& \textbf{28.39} {\color{DarkGreen} (\textbf{-0.39\%})}  & \textbf{0.901} & \textbf{0.159}\\
			\midrule
			\multirow{3}{*}{Noise on Images}
			& MVSNeRF~\citep{chen2021mvsnerf}
			& 24.56 {\color{DarkGreen} (-10.50\%)} & 0.825 & 0.096  \\
			%& OPO-NeuRay & 26.51 {\color{DarkGreen} (\textbf{-3.88\%})} & & \\
			%& OPO-MVSNeRF  & 25.63 {\color{DarkGreen} (-8.17\%)} & 0.854 & 0.085 \\
			& NeuRay~\citep{liu2022neuray} & 25.53 {\color{DarkGreen} (-8.66\%)} & 0.848 & \\
			& OPONeRF  
			& \textbf{27.05} {\color{DarkGreen} (\textbf{-5.09\%})} & \textbf{0.886} & \textbf{0.175}\\
			\bottomrule
		\end{tabular}
	}
	
	\vspace{1pt}
	
	\resizebox{.47\textwidth}{!}{
		\begin{tabular}{clHlH}
			\toprule
			Settings & Method & PSNR$\uparrow$ & SSIM$\uparrow$ & LPIPS$\downarrow$\\
			\midrule
			% \multirow{2}{*}{Original}
			\multirow{3}{*}{Noise Free} & MVSNeRF~\citep{chen2021mvsnerf} & 27.44 &  0.881 & 0.075  \\
			%& OPO-NeuRay & 27.58 & 0.884 &  \\
			%& OPO-MVSNeRF & 27.91 & 0.885 & 0.070 \\ 
			& NeuRay~\citep{liu2022neuray} & 27.94 & 0.888 & \\
			& OPONeRF 
			& \textbf{28.50} & \textbf{0.902} & \textbf{0.158}\\ \midrule
			\multirow{3}{*}{Noise on Features}
			& MVSNeRF~\citep{chen2021mvsnerf}
			% & 24.45 & 0.758 & 0.146  \\
			& 22.36 & 0.623 {\color{DarkGreen} (-29.28\%)} & 0.270  \\
			%& OPO-NeuRay & 27.53 & 0.883 {\color{DarkGreen} (\textbf{-0.11\%})} &  \\
			%& OPO-MVSNeRF & 26.66 & 0.833 {\color{DarkGreen} (-5.88\%)} & 0.098 \\
			& NeuRay~\citep{liu2022neuray} & 27.71 & 0.884 {\color{DarkGreen} (-0.45\%)} & \\
			& OPONeRF 
			& \textbf{28.39} & \textbf{0.901} {\color{DarkGreen} (\textbf{-0.11\%})} & \textbf{0.159}\\
			\midrule
			\multirow{3}{*}{Noise on Images}
			& MVSNeRF~\citep{chen2021mvsnerf}
			& 24.56 & 0.825 {\color{DarkGreen} (-6.36\%)} & 0.096  \\
			%& OPO-NeuRay & 26.51 & 0.871 {\color{DarkGreen} (\textbf{-1.47\%})} &  \\
			%& OPO-MVSNeRF  & 25.63 & 0.854 {\color{DarkGreen} (-3.50\%)} & 0.085 \\
			& NeuRay~\citep{liu2022neuray} & 25.53 & 0.848 {\color{DarkGreen} (-4.50\%)} & \\
			& OPONeRF  
			& \textbf{27.05} & \textbf{0.886} {\color{DarkGreen} (\textbf{-1.77\%})} & \textbf{0.175}\\
			\bottomrule
		\end{tabular}
	}
	\label{tab:results_noise}
\end{table}
	
	\subsubsection{Experimental Results and Analysis}
	The quantitative results are shown in Table~\ref{tab:results_noise}. 
	Both image-level and feature-level cause sizable performance degradation to MVSNeRF. 
	NeuRay achieves favorable robustness against feature-level noises while OPONeRF even further eases the negative effects of test-time noises. 
	On both metrics of PSNR and SSIM, OPONeRF achieves the best results among all the compared trials, which indicate that the flexibility brought by the One-Point-One-NeRF mechanism does not sacrifice the robustness to outlier noises. 
	%not only outperforms its raw implementation on conventional per-scene optimization, but also
	Notably, under test-time feature-level noises, OPONeRF impressively drops by less than $0.4\%$ on both PSNR and SSIM, where the noises can hardly be noticed as shown in Fig.~\ref{fig:teaser}(b). 
	%Apart from the example in Fig.~\ref{fig:teaser}(b), we further show qualitative results on the robustness to test-time noises in the supplementary pages. 

		\subsection{Experiments on Generalization Benchmarks}
		\label{sec:popular_generalization_benchmarks}
		As indicated by Fig.~\ref{fig:setting_comparison}, our proposed experimental setting can be viewed as an extreme case of the conventional generalization-based setting. 
		To validate the effectiveness of OPONeRF on existing popular benchmarks, we further conducted experiments on generalization-based benchmarks.

		\subsubsection{Spatial Reconstruction}
		We firstly validate OPONeRF on the popular generalization benchmarks on spatial reconstruction following \citep{yu2021pixelnerf,trevithick2021grf,chen2021mvsnerf,liu2022neuray}.

		\textbf{Datasets and Settings: }
		Exactly following \citep{liu2022neuray,wang2022attention}, the joint training data consists of both synthetic and real data. 
		For synthetic data, we used the early 1023 object items in the \textbf{Google Scanned Objects}~\citep{downs2022google} dataset. 
		The real data involves all objects in the \textbf{DTU}~\citep{jensen2014large} dataset but ``birds'', ``bricks’', ``snowman’' and ``tools'', the training split of the \textbf{LLFF}~\citep{mildenhall2019local} and \textbf{RealEstate10K}~\citep{zhou2018stereo} datasets
		%\footnote{Due to network error, we only obtained 198 scene instances of the RealEstate10K training set. }
		, 100 objects in the \textbf{Spaces} dataset~\citep{flynn2019deepview}. 
		We adopted the synthetic \textbf{NeRF Synthetic}~\citep{mildenhall2020nerf} dataset and test split of realistic \textbf{DTU}~\citep{jensen2014large} and \textbf{LLFF}~\citep{mildenhall2019local} as test set. 
		
		\textbf{Experimental Results and Analysis: }
		Quantitative results on the generalization benchmarks for spatial reconstruction are summarized in Table~\ref{tab:results_generalization_spatial}. 
		For fair comparison, the compared numbers are drawn from \citep{liu2022neuray,wang2022attention} only while other baselines such as GeoNeRF~\citep{johari2022geonerf} and LIRF~\citep{huang2023local} are trained on different databases. 
		It can be seen that OPONeRF still achieve competitive results compared with advanced generalizable NeRFs such as GNT~\citep{wang2022attention} and GPNR~\citep{suhail2022generalizable}. 
		Especially, it shows the best rendering results when measured by PSNR on all 3 test sets.

		\subsubsection{Spatio-temporal Reconstruction}
		Then we experimented on the spatio-temporal reconstruction to test the generalization to unseen motions following \citep{tian2023mononerf}. 
		
		\textbf{Datasets and Settings: } The employed database contains the ``Balloon2'' and ``Truck'' scenes in the \textbf{Dynamic Scene}~\citep{gao2021dynamic} dataset. 
		The dataset contains video sequences captured by a 12-camera rig. 
		From 2 scenes, we extract the same 12 frames with \citep{tian2023mononerf}, each captured by a unique camera at a unique timestamp to simulate a moving camera shooting a monocular video. 
		Following \citep{tian2023mononerf}, we used the first 4 timeframes for training and last 8 with unseen motions for testing. 
		Since OPONeRF is not specially intended for spatio-temporal reconstruction, we instead incorporated OPONeRF into MonoNeRF, following its official settings on hyperparameters and training protocols. 
		Target layers are restricted within the dynamic branch of MonoNeRF~\citep{tian2023mononerf}. 
		%Please refer to subsection~ for the details of One-Point-One-MonoNeRF (OPO-MonoNeRF). 
		\begin{table}[bt!]
			%\small
			\caption{Quantitative comparison with baseline methods on the generalization benchmarks for spatial reconstruction. 
				Highlights are the \textbf{best} and \underline{runner-up}. 
			}
			\label{tab:results_generalization_spatial}
			\vspace{-2pt}
			\centering
			\resizebox{.48\textwidth}{!}{
				\begin{tabular}{lcHcccHcccHc}
					\toprule
					& \multicolumn{3}{c}{NeRF Synthetic} & & \multicolumn{3}{c}{Real Object DTU} & & \multicolumn{3}{c}{LLFF} \\
					Method & PSNR$\uparrow$ & SSIM$\uparrow$ & LPIPS$\downarrow$ & & PSNR$\uparrow$ & SSIM$\uparrow$ & LPIPS$\downarrow$ & & PSNR$\uparrow$ & SSIM$\uparrow$ & LPIPS$\downarrow$ \\
					\midrule
					% & LLFF~\citep{mildenhall2019local}
					% & 24.77 & 0.911 & 0.114 &    -  &     - &     - & 24.41 & 0.805 & 0.211 \\
					
					PixelNeRF~\citep{yu2021pixelnerf} 
					& 22.65 & 0.808 & 0.202 & & 19.40 & 0.463 & 0.447 & & 18.66 & 0.588 & 0.463 \\
					MVSNeRF~\citep{chen2021mvsnerf}
					& 25.15 & 0.853 & 0.159 & & 23.83 & 0.723 & 0.286 & & 21.18 & 0.691 & 0.301 \\
					IBRNet~\citep{wang2021ibrnet}
					& 26.73 & 0.908 & 0.101 & & 25.76 & 0.861 & 0.173 & & 25.17 & 0.813 & 0.200 \\
					% & Ours      & \textbf{28.91} & \textbf{0.920} & \textbf{0.095} & \textbf{28.30} & \textbf{0.907} & \textbf{0.130} & \textbf{25.85} & \textbf{0.832} & \textbf{0.190} \\
					GPNR~\citep{suhail2022generalizable} & 26.48 & 0.944 & \textbf{0.055} & & -- & -- & -- & & 25.72 & 0.880 & \underline{0.175} \\
					GNT~\citep{wang2022attention} & 27.29 & \textbf{0.937} & \underline{0.056} & & -- & -- & -- & & \underline{25.86} & \textbf{0.867} & \textbf{0.116}\\
					NeuRay~\citep{liu2022neuray} & \underline{28.29} & \underline{0.927} & 0.080 & & \underline{26.47} & \underline{0.875} & \underline{0.158} & & 25.35 & 0.818 & 0.198 \\
					%NeuRay$^{*}$~\citep{liu2022neuray}  & \underline{27.42} & 0.899 & 0.107 &  \underline{27.31} & \textbf{0.901} & \textbf{0.127} & 24.76 & 0.809 & \underline{0.198}  \\
					%NeuRay$^{\}$~\citep{liu2022neuray} & 27.20 & 0.889 & 0.115 & 26.65 & 0.879 & 0.151 & \underline{24.86} & \underline{0.810} & 0.201 \\ 
						\midrule
						OPONeRF & \textbf{28.59} & 0.912 & 0.096 & & \textbf{27.45} & \textbf{0.900} & \textbf{0.140} & & \textbf{25.91} & \underline{0.832} & 0.181\\
						%\multirow{4}{*}{Finetuning}	& MVSNeRF~\citep{chen2021mvsnerf} &  27.21 & 0.888 & 0.162 & 25.41 & 0.767 & 0.275 & 23.54 & 0.733 & 0.317 \\
						% & NeRF~\citep{mildenhall2020nerf} & 31.01 & 0.947 & 0.081 & 28.11 & 0.860 & 0.207 & 26.74 & 0.840 & 0.178 \\
						%& IBRNet~\citep{wang2021ibrnet}	& 30.05 & 0.935 & 0.066 & 29.17 & 0.908 & 0.128 & 26.87 & 0.848 & 0.175  \\
						% & NeuRay~\citep{liu2022neuray}   & \textbf{32.35} & \textbf{0.960} & \textbf{0.048} & \textbf{29.79}  & \textbf{0.928} & \textbf{0.107} &  \textbf{27.06} & \textbf{0.850} & \textbf{0.172} \\
						\bottomrule
				\end{tabular}}
			\end{table}
			\begin{table}[bt!]
				%\small
				\caption{Quantitative comparison with baseline methods on the generalization benchmarks for spatio-temporal reconstruction. 
					%Highlights are the \textbf{best} and \underline{runner-up}. %$^{*}$ denotes that the results are reproduced from the pretrained model provided by \citep{liu2022neuray}. $^{\dagger}$ indicates our reproduced results following the official settings in \citep{liu2022neuray}. 
				}
				\label{tab:results_generalization_spatialtemporal}
				\vspace{-2pt}
				\centering
				\resizebox{.48\textwidth}{!}{
					\begin{tabular}{lcccHccc}
						\toprule
						& \multicolumn{3}{c}{Balloon2} & & \multicolumn{3}{c}{Truck} \\
						Method & PSNR$\uparrow$ & SSIM$\uparrow$ & LPIPS$\downarrow$ & & PSNR$\uparrow$ & SSIM$\uparrow$ & LPIPS$\downarrow$ \\ \midrule
						NeRF~\citep{mildenhall2020nerf} & 20.33 & 0.662 & 0.256 & & 20.26 & 0.669 & 0.224\\ 
						NeRF~\citep{mildenhall2020nerf} + time & 20.22 & 0.661 & 0.218 & & 20.26 & 0.639 &   0.256\\
						DynNeRF~\citep{gao2021dynamic} & 19.99 & 0.641 & 0.291 & & 20.33 & 0.621 & 0.273\\
						MonoNeRF~\citep{tian2023mononerf} & 21.30 & 0.669 & 0.204 & & 23.74 & 0.702 & 0.174\\
						%MonoNeRF$^{*}$~\citep{tian2023mononerf} & \underline{23.34} & \underline{0.748} & \underline{0.119} & & \underline{24.29} & \underline{0.719} &  \underline{0.160} \\
						\midrule
						OPO-MonoNeRF & \textbf{23.52} & \textbf{0.756} & \textbf{0.116} & & \textbf{24.71} & \textbf{0.732} & \textbf{0.151} \\ 
						\bottomrule
				\end{tabular}}	
			\end{table}
			\begin{figure}[tb]
				\centering
				\includegraphics[width=0.48\textwidth]{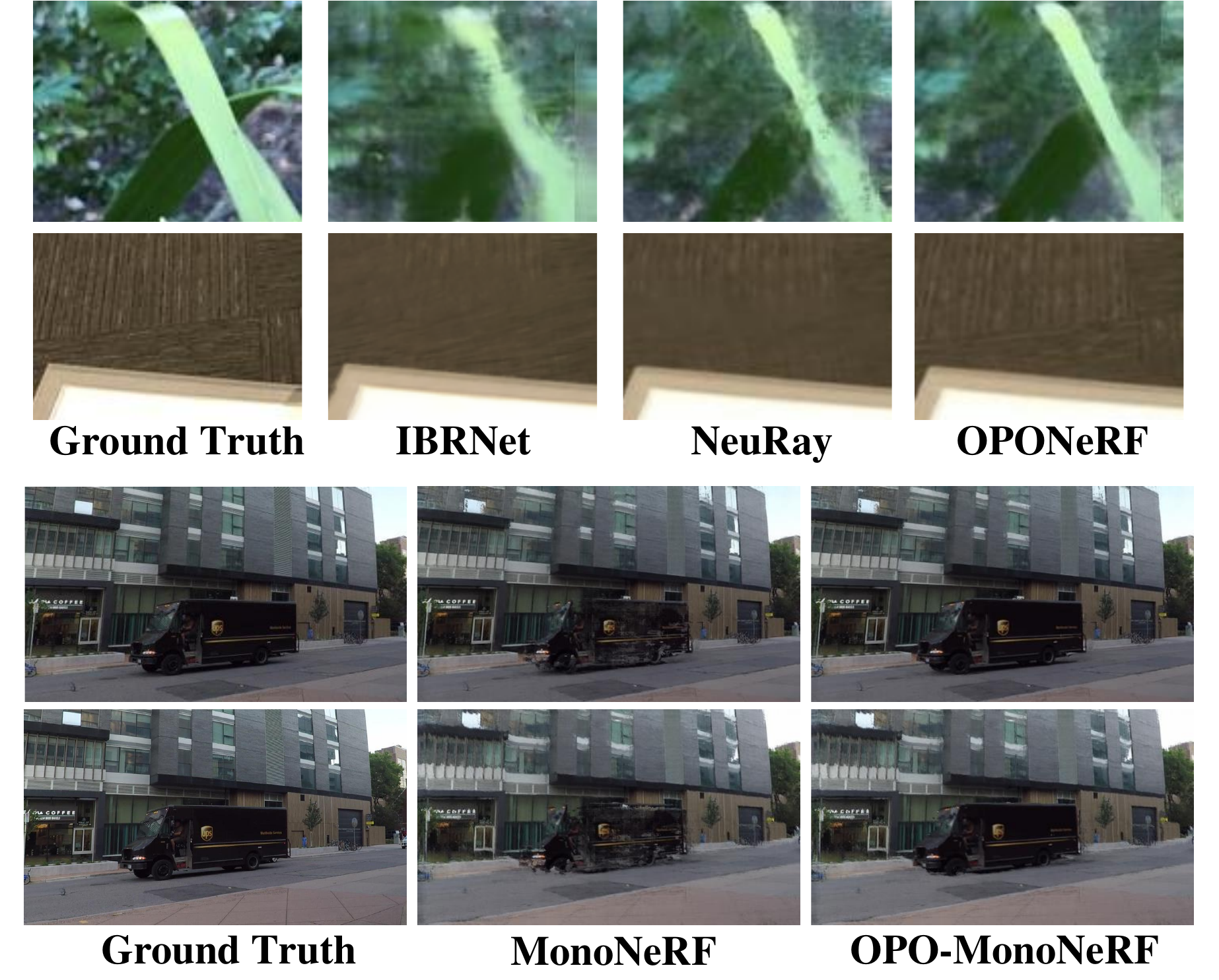}
				\vspace{-6mm}
				\caption{
					%	Qualitative comparison between state-of-the-art methods~\citep{wang2021ibrnet,liu2022neuray} and OPONeRF on LLFF. 
					%Qualitative comparison on LLFF. 
					Qualitative results on LLFF for spatial reconstruction and Dynamic Scene for spatio-temporal reconstruction. 
					%	(\textit{Best viewed when zoomed in.})
				} \label{fig:generalization_qualitative}
			\end{figure}
		
		\textbf{Experimental Results and Analysis: }
		Quantitative results on the generalization benchmarks for spatio-temporal reconstruction are summarized in Table~\ref{tab:results_generalization_spatialtemporal}. 
		%The best and runner-up numbers are highlighted. 
		Our MonoNeRF implementation consistently improves the generalization to unseen motions upon the MonoNeRF baseline. 
		Given better expressiveness and stronger flexibility, it can better model the sophisticated physical motions which widely exist in real world. 
		
		Fig.~\ref{fig:generalization_qualitative} shows the qualitative results on the generalization benchmarks for spatial and spatio-temporal reconstruction. 
		OPONeRF improves the detail preserving of unseen scenes or motions via the comparison with other baseline methods. 
		Quantitatively, the improvement of OPONeRF is much more obvious in spatio-temporal reconstruction. 
		We infer that our OPONeRF helps to significantly fertile the training samples from an ill-posed monocular video. 
		In contrast, traditional generalization-based benchmarks for spatial reconstruction are provided with sufficient training views, which benefits the network design with stronger scene priors.

		\subsection{Ablation Studies and Discussions}
		\label{sec:discussion_and_ablation}
		In this subsection, we conducted ablation studies on the efficacy of the components in OPONeRF and choices of hyper-parameters. 
		We then discuss the computational efficiency of OPONeRF and the comparison with spatio-temporal reconstruction benchmark. 
		We also demonstrate the general applicability of OPONeRF via incorporating it into other advanced baseline methods. 
		
		\textbf{Efficacy of Designed Components: }
		The trial of removing adaptiveness factor $\mathbf{a}_x$ is indicated by ID 1 in Table~\ref{tab:results_ablation_component}, where the responses to the test-time scene changes are solely guided by the shared geometric representation and probabilistic point representation. 
		Compared with default implementation of OPONeRF, the trial achieves lower PSNR but still higher than NeuRay. 
		This indicates that the adaptiveness to the given scene is truly related to spatial locality and requires discriminative learning w.r.t. sampled coordinates, which is effectively captured by our proposed adaptiveness factor. 
		ID 2 represents using vanilla $f_\mathbf{x}$ and $a_\mathbf{x}$ and removing the usage of probabilistic inference, i.e., $\mathcal{L}_{rec}$ and $\mathcal{L}_{appr}$.  
		The rendering quality drop by 1.50 PSNR and 1.10 PSNR on 2 scenes respectively, which indicates that OPONeRF can effectively respond to test-time scene changes via the unsupervised modeling of unexpected local variance. 
		
		\begin{table}
			\centering
			\caption{
				Experimental results of ablation studies on OPONeRF. 
				``Beef'' and ``White'' indicates the ``Cut Roasted Beef'' task of realistic benchmark and scene set with white background of synthetic benchmark. 
				The metric is PSNR ($\uparrow$). 
			}
			\vspace{-2pt}
			\resizebox{.46\textwidth}{!}{
				\begin{tabular}{cccc}
					\toprule
					Trial ID & Description & Beef & White\\ \midrule
					-- & OPONeRF & 31.76 & 35.23 \\
					-- & NeuRay & 28.86 & 33.52 \\ \midrule
					1 & W/o adaptiveness factor & 29.32 &  33.87 \\ 
					2 & W/o probabilistic modeling & 30.26 & 34.13\\ \midrule
					3 & Directly regression & 29.51 & 32.08 \\
					4 & Interpolation $\rightarrow$ fusion in Eqn.~(\ref{eq:adaptiveness_factor_fusion}) & 31.76 & 35.23 \\ \midrule
					5 & All ones for $\mathcal{}{A}_\mathbf{x}$ in Eqn.~(\ref{eq:layer_mask}) & 28.93 & 33.52 \\
					6 & All zeros for $M^l$ in Eqn.~(\ref{eq:soft_mask_generation}) & 24.05 & 27.89  \\
					7 & W/o quantization in PCD module & 24.20 & 30.79 \\ 
					8 & W/o diversity loss & 31.45 & 34.43 \\ \midrule
					9 & W/o residual connection in $\mathcal{F}_\mathbf{x}$ & 26.41 & 28.10\\
					10 & W/o $f_\mathbf{x}^I$ in $\mathcal{F}_\mathbf{x}$& 27.00 & 29.25\\ 
					%8 & w/o Diversity Loss &  &  \\
					\bottomrule
				\end{tabular}
			}
			\label{tab:results_ablation_component}
		\end{table}
		\begin{figure}[tb]
			\centering
			\includegraphics[width=0.42\textwidth]{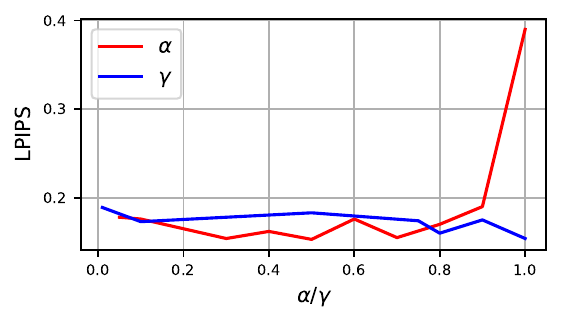}
			\vspace{-4mm}
			\caption{
				%	Qualitative comparison between state-of-the-art methods~\citep{wang2021ibrnet,liu2022neuray} and OPONeRF on LLFF. 
				%Qualitative comparison on LLFF. 
				LPIPS($\downarrow$) values after ablating $\alpha$ and $\gamma$. For the $\alpha$-curve, $\gamma$ is set as 0.3. For the $\gamma$-curve, $\alpha$ is set as 1.0. 
				%	(\textit{Best viewed when zoomed in.})
			} \label{fig:loss_ablation_study}
		\end{figure}
		
		\textbf{Way of Obtaining Adaptiveness Factor: }
		By default we firstly interpolate using spatial query then apply regression layer to obtain the adaptiveness factor $\mathbf{a}_x$ (ID 3 in Table~\ref{tab:results_ablation_component}). 
		An alternative is to keep the fan-out dimension unchanged and directly regress $\mathcal{F}_\mathbf{x}$ into $\mathcal{A}_\mathbf{x}$ and discard the fusion in Eqn.~(\ref{eq:adaptiveness_factor_fusion}) (ID 4 in Table~\ref{tab:results_ablation_component}). 
		This implementation shows inferior performance which implies the importance of learning a geometric-aware adaptiveness factor. 
		Also, the default implementation models $f_\mathbf{x}$ and $a_\mathbf{x}$ separately which tackles the uncertainty in scene representation and the corresponding responses to scene variations respectively. 
		Such a design choice outperforms the naive mapping from point to NeRF as indicated by the quantitative results. 
		
		\textbf{Flexibility of Learned Adaptiveness: }
		The flexibility of learned adaptiveness is largely controlled by the binary weight mask $M^l$ and soft layer mask $\mathcal{A}_\mathbf{x}$. 
		Here we ablated such flexibility by setting them as all-ones and all-zeros matrices respectively (ID 5 and 6 in Table~\ref{tab:results_ablation_component}). 
		Note that ID 5 equals discarding all divide-and-conquer designs, which suffers from sensitivity to scene changes. 
		The poorer results of trial ID 6 indicates the strategy of learning a set of layer-variant embeddings for candidate weights can effectively prevent performance degradation. 
		After removing the quantization in the PCD module and use soft $M^l$ (ID 7 in Table~\ref{tab:results_ablation_component}), significant performance drop is observed. 
		This implies the rich geometric features condensed in parameter candidates can be better preserved via directly retrieving their individual elements. 
		Then we tested the removal of $\mathcal{L}_{div}$ (ID 8). 
		The usage of $\mathcal{L}_{div}$ strengthens the diverse response of OPONeRF to local variations, especially in synthetic database which shows larger performance decline after removing $\mathcal{L}_{div}$ ($\downarrow$0.80 v.s. $\downarrow$0.31).

		\textbf{Decomposition of Point Representation: }
		By default we represent the final $\mathcal{F}_\mathbf{x}$ as the combination of deterministic invariance ($f_\mathbf{x}^I$) and unexpected variance ($f_\mathbf{x}^V$), along with a residual connection ($f_\mathbf{x}$). 
		Here we ablated the component of $\mathcal{F}_\mathbf{x}$ via removing $f_\mathbf{x}^I$ or $f_\mathbf{x}$, indicated by ID 9 and 10 respectively in Table~\ref{tab:results_ablation_component}. 
		We observe that the removal of residual connection or deterministic invariance causes enormous rendering degradation, of which the trial 9 presents a severer PSNR decline. 
		We infer that while $f_\mathbf{x}^I$ and $f_\mathbf{x}$ are both supposed to model the deterministic mapping from the queried point to the renderer parameters, $f_\mathbf{x}$ further helps with the steady gradients as the residual path~\citep{he2016deep}. 
		
		\textbf{Values of Weight Hyper-parameters: }
		The hyper-parameters $\alpha$ in $\mathcal{F}_\mathbf{x}$ and $\gamma$ in $\mathcal{F}_{opo}$ control the gradients of the final point representation and the balance between $\mathcal{L}_{appr}$ and $\mathcal{L}_{rec}$. 
		We tested different values and have summarized the results on ``Beef'' scene in Fig.~\ref{fig:loss_ablation_study}. 
		We can see that the rendering with test-time changes is more sensitive when tuning the ingredients of $\mathcal{F}_\mathbf{x}$. 
		%, which suggests the necessity of residual connection in $\mathcal{F}_\mathbf{x}$. 
		As for the balance between $\mathcal{L}_{appr}$ and $\mathcal{L}_{rec}$, we observe minor fluctuations when ablating $\gamma$.  
		By default we set $\alpha=0.3$ and $\gamma=1.0$. 
		
		\textbf{Computational Efficiency: }
		We counted the total FLOPS when rendering a image with the resolution of 2704x2028, and compared with NeuRay which showed the best results among all the baseline methods. NeuRay and OPONeRF cost 45.3 and 46.2 GFLOPS respectively. 
		The corresponding throughput is 10.7 and 11.1 seconds per frame on a single RTX3090 GPU. 
		
		\textbf{Infeasibility of Spatio-temporal Construction in Our Benchmark: }
		To demonstrate the difficulties in spatio-temporal reconstruction in our settings where only a single timeframe is provided, we conducted experiments on our Cooking-Perturbation benchmark using D-NeRF and compared with OPONeRF. 
		We experimented on the ``Cut Roasted Beef'', ``Flame Steak'' and ``Sear Steak'' tasks. 
		Quantitative results in Table.~\ref{tab:comparison_dnerf} validate that D-NeRF performs poorly in the absence of temporal variations, where the rendering quality of OPONeRF is improved significantly measured by LPIPS. 
		
		\begin{table}[bt!]
			\centering
			%\footnotesize
			\caption{
				Quantitative comparison between D-NeRF and OPONeRF on our Cooking-Perturbation benchmark, evaluated by mean LPIPS$\downarrow$. 
				The assumptions in D-NeRF do not hold in our experimental setting. 
			}
			\vspace{-2pt}
			\begin{adjustbox}{width=.44\textwidth,center}
				\begin{tabular}{l|lll}
					\toprule
					& Cut Roasted Beef & Flame Steak & Sear Steak \\ \hline
					D-NeRF~\citep{pumarola2021d} & 0.459 & 0.536 & 0.482\\
					OPONeRF & \textbf{0.149}  {\color{red} (-67.54\%)}  & \textbf{0.116} {\color{red} (-78.36\%)} & \textbf{0.119} {\color{red} (-75.31\%)} \\
					%Dynamic & Small & Small & D-NeRF, DynNeRF, etc. \\
					%Vanilla & None & None & NeRF, RegNeRF, etc. \\
					\bottomrule
				\end{tabular}
			\end{adjustbox}
			\label{tab:comparison_dnerf}
		\end{table}
		
		\begin{table}[bt!]
			\caption{Quantitative comparison with baseline methods with and without incorporating OPONeRF into their renderer architectures. 
				We did not test GeoNeRF or OPO-GeoNeRF on the ``VR Headset'' task because they both produce all-black results which are totally indistinguishable. 
			}
			\vspace{-2pt}
			\centering
			\resizebox{.48\textwidth}{!}{
				\begin{tabular}{llll}
					\toprule
					& \multicolumn{3}{c}{Sear Steak}\\
					Method & PSNR$\uparrow$ & SSIM$\uparrow$ & LPIPS$\downarrow$\\
					\midrule
					% & ~\citep{mildenhall2019local}
					% & 24.77 & 0.911 & 0.114 &    -  &     - &     - & 24.41 & 0.805 & 0.211 \\
					GeoNeRF~\citep{johari2022geonerf} 
					& 17.05 & 0.408 & 0.490
					\\
					OPO-GeoNeRF & \textbf{18.78} {\color{red} (+1.73)} & \textbf{0.419} {\color{red} (+0.011)} &  \textbf{0.273} {\color{red} (-0.217)}\\ \midrule
					MVSNeRF~\citep{chen2021mvsnerf}
					& 23.87 & 0.586 & 0.256\\
					OPO-MVSNeRF
					& \textbf{25.13} {\color{red} (+1.26)} & \textbf{0.680} {\color{red} (+0.094)} & \textbf{0.204} {\color{red} (-0.052)} \\ \midrule
					%IBRNet~\citep{wang2021ibrnet} & 26.39 & 0.803 & 0.238 & 24.27 & 0.709 & 0.182 & 25.71 & 0.762 & 0.225 & 23.86 & 0.747 & 0.276 \\
					NeuRay~\citep{liu2022neuray}
					& 30.95 & 0.937 & 0.122 \\ 
					OPO-NeuRay
					& \textbf{31.19} {\color{red} (+0.24)} & \textbf{0.939} {\color{red} (+0.002)} & \textbf{0.119} {\color{red} (-0.003)} \\
					\bottomrule
				\end{tabular}
			}
			\newline
			\vspace{2pt}
			\newline
			\resizebox{.48\textwidth}{!}{
				\begin{tabular}{llll}
					\toprule
					& \multicolumn{3}{c}{VR Headset} \\
					Method & PSNR$\uparrow$ & SSIM$\uparrow$ & LPIPS$\downarrow$\\
					\midrule
					% & ~\citep{mildenhall2019local}
					% & 24.77 & 0.911 & 0.114 &    -  &     - &     - & 24.41 & 0.805 & 0.211 \\
					%\rowcolor{Gray}
					%GeoNeRF~\citep{johari2022geonerf} & 6.95 & 0.003 & 1.007 \\
					MVSNeRF~\citep{chen2021mvsnerf}
					& 12.39 & 0.443 & 0.438\\
					OPO-MVSNeRF
					& \textbf{12.61} {\color{red} (+0.22)} & \textbf{0.442} {\color{DarkGreen} (-0.001)} & \textbf{0.458} {\color{DarkGreen} (+0.020)} \\ \midrule
					%IBRNet~\citep{wang2021ibrnet} & 26.39 & 0.803 & 0.238 & 24.27 & 0.709 & 0.182 & 25.71 & 0.762 & 0.225 & 23.86 & 0.747 & 0.276 \\
					NeuRay~\citep{liu2022neuray}
					& 28.61 & 0.937 & 0.153
					\\ 
					%OPO-GeoNeRF &  &  &  \\
					
					OPO-NeuRay & \textbf{29.35} {\color{red} (+0.74)} & \textbf{0.940} {\color{red} (+0.003)} & \textbf{0.146} {\color{red} (-0.007)} \\
					\bottomrule
				\end{tabular}
			}
			\label{tab:results_baseline_incorporation}
		\end{table}
		
		\textbf{Efficacy of Incorporating OPONeRF into Other Baselines: }
		The validate the general applicability of OPONeRF, we have incorporated it into popular baseline methods such as MVSNeRF~\citep{chen2021mvsnerf}, NeuRay~\citep{liu2022neuray} and GeoNeRF~\citep{johari2022geonerf}. 
		We conducted experiments on the ``Sear Steak'' scene in the Cooking-Perturbation benchmark and ``VR Headset'' scene in the MeetRoom-Perturbation benchmark. 
		Results of OPO-GeoNeRF and GeoNeRF on ``VR Headset'' are not reported because they produce all-black results, which also indicate the infeasibility of GeoNeRF (and its variant) on our constructed benchmark despite its powerful generalization ability. 
		The improvements on the ``Sear Steak'' scene are consistently more significant than ``VR Headset''. 
		We infer that it is due to the more complicated scene layout in ``Sear Steak'' which contributes to more abundant local variations. 
		
		\begin{figure}[bt!]
			\centering
			\includegraphics[width=0.44\textwidth]{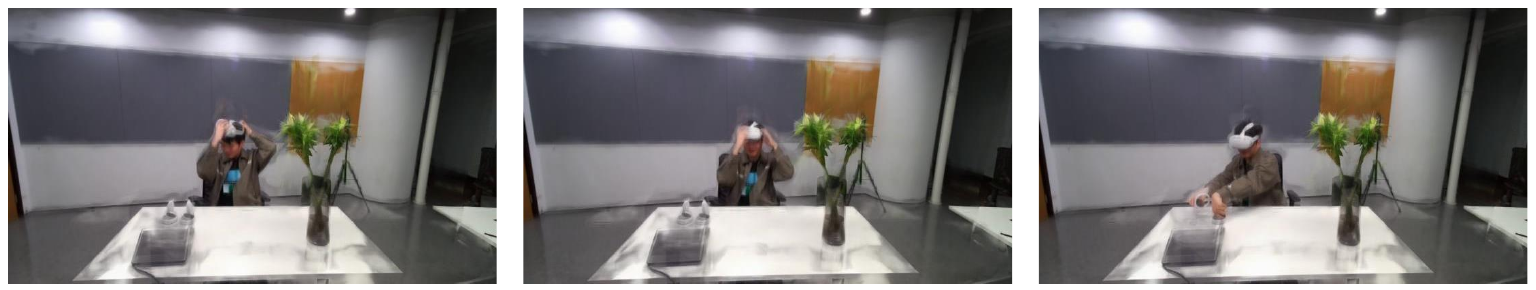}
			\vspace{-2mm}
			\caption{
				%	Qualitative comparison between state-of-the-art methods~\citep{wang2021ibrnet,liu2022neuray} and OPONeRF on LLFF. 
				%Qualitative comparison on LLFF. 
				Failure cases by OPONeRF (trained on Spinach). 
				%	(\textit{Best viewed when zoomed in.})
			} \label{fig:failure_case}
		\end{figure}
		
		\textbf{Limitations of OPONeRF: }
		Given one training timestamp and not specifically designed for scenes with variable daylight~\citep{song2019neural}, OPONeRF struggles at handling complex illuminations. 
		As indicated by Fig.~\ref{fig:failure_case}, OPONeRF also produces blurred results when faced with significant scene changes, e.g., trained on Cook Spinache and evaluated on VR Headset. 
		Moreover, compared with the recent Gaussian Splatting method~\citep{wu20244dgaussians}, OPONeRF shares with other NeRF-based methods the slower training speed. 
\section{Conclusion}
In this paper, we propose a One-Point-One NeRF (OPONeRF) for robust scene rendering. 
We design a divide-and-conquer framework which adaptively responds to local scene variations via personalizing proper point-wise renderer parameters. 
To model the uncertainty explicitly, we further decompose the point representation into deterministic invariance and probabilistic variance. 
We have also built benchmarks with diverse test-time scene changes to validate the proposed method. 
Extensive experiments on our constructed databases and existing popular benchmarks validate that OPONeRF presents superb rendering performances compared with other methods with advanced generalization ability. 
We also demonstrate the general applicability of OPONeRF via the incorporation into existing popular baselines. 
Further works include the validation of our approaches on the unbounded and large-scale urban databases. 
% which might be released in future. 

\section*{Acknowledgements}
This work was supported in part by the National Key Research and Development Program of China under Grant 2022ZD0160102, and in part by the National Natural Science Foundation of China under Grant 62125603, Grant 62206147, Grant 62321005, and Grant 62336004. 

\section*{Data availability} The data in Cooking-Perturbation and MeetRoom-Perturbation benchmarks are re-organized from the publicly available datasets released on \url{https://github.com/facebookresearch/Neural_3D_Video} and \url{https://github.com/AlgoHunt/StreamRF} respectively. 
The dataset for simulating illumination variation is synthesized by using the Kubric dataset engine on \url{https://github.com/google-research/kubric}. 
The Google Scanned Objects, DTU, LLFF, RealEstate10K, Spaces, NeRF Synthetic datasets for generalizable spatial reconstruction are available on \url{https://github.com/googleinterns/IBRNet} and \url{https://github.com/liuyuan-pal/NeuRay}. 
The Dynamic Scene dataset for generalizable spatio-temporal reconstruction can be accessed at \url{https://github.com/tianfr/MonoNeRF}.

\section*{Compliance with Ethical Standards}
\textbf{Conflict of interest }
The authors declare that they have no conflict of interest. 

\noindent\textbf{Ethical approval }
This article does not contain any studies with human participants or animals. 
\\

\iffalse
\begin{appendices}
\renewcommand{\thesection}{\Alph{section}.}
\noindent\textbf{Appendix}
\section{Experiment on Test-time Object Adding or Removal}
\end{appendices}
\fi
\bibliographystyle{spbasic}      % basic style, author-year citations
\bibliography{egbib}   % name your BibTeX data base

\end{document}